%% file: main.tex
\documentclass{article}

\usepackage[english]{babel}

\usepackage[letterpaper,top=2cm,bottom=2cm,left=3cm,right=3cm,marginparwidth=1.75cm]{geometry}

\usepackage[colorlinks=true, allcolors=blue]{hyperref}

\usepackage[utf8]{inputenc} 
\usepackage[T1]{fontenc}    
\usepackage{url}            
\usepackage{booktabs}       
\usepackage{amsfonts}       
\usepackage{nicefrac}       
\usepackage{microtype}      
\usepackage{xcolor}         

\usepackage{graphicx}
\usepackage{subcaption}
\usepackage{multirow}

\usepackage{amsmath}
\usepackage{amssymb}
\usepackage{mathtools}
\usepackage{amsthm}
\usepackage{makecell}
\usepackage{cuted}
\usepackage{stfloats}
\usepackage{algorithm}
\usepackage{algpseudocode}
\usepackage{wrapfig}
\usepackage{natbib}
\usepackage{array}
\usepackage[capitalize,noabbrev]{cleveref}

\usepackage{tabularx}

\input{our_commands.tex}

\theoremstyle{plain}

\theoremstyle{definition}

\theoremstyle{remark}

\usepackage[textsize=tiny]{todonotes}
\usepackage[hang,flushmargin]{footmisc}

\title{
OrbiSim: World Models as Differentiable Physics Engines for Embodied Intelligence
\footnotetext{\textit{Preprint.} This work is part of the NeoWorld Project.}}
\author{Jiajian Li$^{1,2}$\thanks{Equal contribution.}
\qquad
Jingyuan Huang$^{1,2}$$^*$
\qquad
Junru Gong$^{1,2}$$^*$
\\ 
Qi Wang$^{1}$
\qquad
Xiaokang Yang$^{1}$
\qquad
Yunbo Wang$^{1,2}$\thanks{Corresponding author: Yunbo~Wang <yunbow@sjtu.edu.cn>.}\vspace{5pt}
\\
$^1$MoE Key Lab of Artificial Intelligence, Shanghai Jiao Tong University\\
$^2$Doxi AI
}

\begin{document}
\maketitle

\begin{abstract}
  We present OrbiSim, a novel robotic simulation paradigm that redefines world models as a fully differentiable physics engine for embodied intelligence.
  Unlike prior world models that focus on unconstrained imagination in latent or visual domains, OrbiSim establishes a unified, physically-grounded pathway that bridges structured scene assets, neural dynamics, and downstream reinforcement learning.
  By enabling end-to-end differentiability throughout the entire simulation loop---spanning from explicit state transitions to visual observation generation---OrbiSim supports tasks traditionally intractable for classical simulators, such as differentiable contact modeling, gradient-based policy optimization under sparse rewards, and intuitive physical inference.
  Empirical results demonstrate that OrbiSim significantly outperforms state-of-the-art world models in both predictive fidelity and control performance. Furthermore, its consistent responsiveness to asset configurations and physical parameters suggests its potential as a differentiable tool for enhancing robot simulation and policy training. To view detailed demonstrations, please visit \href{https://jjleejj85.github.io/projects/orbisim}{our website}.
\end{abstract}

\begin{table}[h]
    \centering
    \small
    \setlength{\tabcolsep}{5pt}
    \caption{\textbf{Comparison of OrbiSim with traditional physics engines and world models.}}
    \label{tab:setting_cmp}
    \begin{tabular}{lccc}
    \toprule
    \multirow{2}{*}{Features}
    & Classical Physics Engines
    & Generative World Models
    & \textbf{OrbiSim} \\
    & \footnotesize (\textit{e.g.}, Isaac Sim~\citep{isaaclab})
    & \footnotesize (\textit{e.g.}, Motus~\cite{bi2025motusunifiedlatentaction})
    & \textbf{\footnotesize (Ours)} \\
    \midrule
    Primary Focus
    & High-Fidelity Simulation
    & Video Synthesis
    & Full-Stack Diff. Sim \\
    
    Core Mechanism
    & Numerical Solvers
    & Neural Networks
    & Decoupled Neural Networks \\
    
    Asset Control
    & \mycheckmark Full
    & \myxmark
    & \mycheckmark Basic Geometry + Physics \\
    
    Differentiability
    & \myxmark
    & \mycheckmark Partial
    & \mycheckmark End-to-End \\
    
    Output Modality
    & State + Rendered Pixels
    & Rendered Pixels
    & State + Rendered Pixels \\
    
    \bottomrule
    \end{tabular}
\end{table}

\section{Introduction}

The advancement of embodied intelligence is intrinsically tied to the fidelity of simulation software~\citep {todorov2012mujoco,robosuite,isaaclab}, with classical physics engines such as MuJoCo~\citep{todorov2012mujoco}, PhysX~\citep{physx}, and Bullet~\citep{bullet} serving as the core execution backends of these platforms.
Despite their success, these numerical engines are fundamentally limited by non-differentiable or partially differentiable simulation pipelines.
More crucially, their hand-crafted numerical solvers impose rigid analytical assumptions, which often lead to a pronounced simulation-to-reality gap, particularly in contact-rich scenarios involving discontinuous dynamics and complex physical interactions.

In parallel, recent years have seen rapid advances in world models that learn predictive representations of environment dynamics directly from data~\citep{bruce2024genie,gao2025adaworld,worldgym2025}. These models have demonstrated impressive capabilities in visual perception, video prediction, and counterfactual reasoning for planning and reinforcement learning (RL). 
Nevertheless, most existing world models act more as visual predictors than as functional simulators, lacking the structural integrity required to replace traditional engines in a closed-loop training pipeline.
%
They typically do not expose explicit physical states, do not accept structured scene assets or physical parameters as inputs, and do not provide a fully differentiable execution pathway that connects environment configuration to downstream control objectives.

\begin{figure*}[t]
    \centering
    \includegraphics[width=\textwidth]{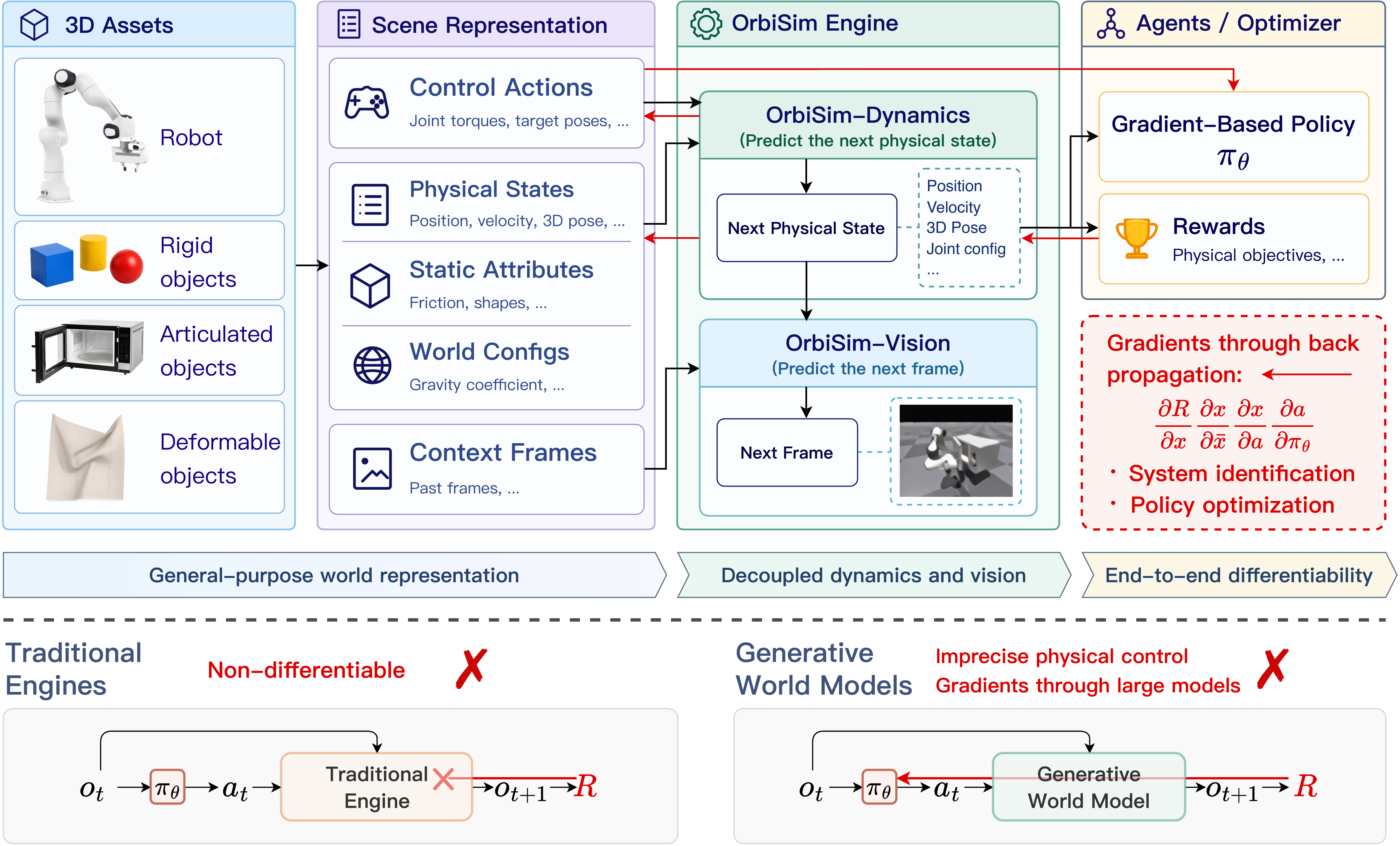}
    \caption{\textbf{Overview of OrbiSim.} Unlike prior generative world models, OrbiSim is designed as a \textit{differentiable physics engine} that seamlessly integrates with modern robotic simulation software, enabling asset control and joint state-pixel generation.
    OrbiSim provides analytical gradients for direct policy optimization, which is a critical capability that distinguishes it from traditional simulators.
    }
    \label{fig:overview}
\end{figure*}

To bridge this gap, we introduce \textit{OrbiSim}, a framework that redefines world modeling as a full-stack, differentiable physics engine for embodied robotics.
Unlike prior architectures that entangle physics with pixels, OrbiSim treats structured geometry and physical properties as first-class conditioning signals (Figure~\ref{fig:overview}), realized through two foundational designs:
First, it incorporates a general-purpose, asset-conditioned interface. Rather than binding to task-specific configurations, \model{} introduces a general-purpose, asset-conditioned interface that unifies heterogeneous physical entities, including rigid bodies, joint-constrained articulated structures, and deformable materials, through modular geometric and physical descriptors.
Second, OrbiSim adopts a decoupled dynamics and vision architecture. \model{} separates physical evolution from visual synthesis by interlinking an object-centric dynamics core (\textit{\model{}-Dynamics}) with a physically grounded latent renderer (\textit{\model{}-Vision}). This separation provides dual state-pixel observation interfaces while maintaining end-to-end differentiability across the entire execution trace.

Compared with conventional physics simulators, \model{} delivers two key advantages:
\begin{itemize}
    \item
    \textit{Data-Grounded Real-to-Sim Alignment:} By integrating structured 3D assets with real-world visual observations, OrbiSim provides a principled pathway toward reducing the simulation-to-reality gap. Rather than relying exclusively on analytical models and idealized physical assumptions, it enables empirical interaction dynamics to be learned directly from observational data.
    \item \textit{End-to-End Differentiability:} OrbiSim enables direct gradient-based policy optimization through its differentiable dynamics core. By bypassing visual synthesis during policy backpropagation, it avoids the noisy gradients and computational overhead associated with differentiating through pixel generation, enabling more efficient optimization and improved learning in sparse-reward settings. Moreover, differentiability supports gradient-based system identification for real-world control, allowing latent physical properties, such as material parameters, to be inferred by minimizing discrepancies between predicted and observed states or visual outcomes.
\end{itemize}



We evaluate OrbiSim across a diverse suite of physics-rich benchmarks.
Our results demonstrate that OrbiSim achieves competitive physical consistency with classical physics solvers while substantially outperforming state-of-the-art world models in video prediction fidelity and long-horizon rollout stability.
Beyond standard benchmarks, we demonstrate OrbiSim's practical utility in three settings: Real-Sim-Real robot manipulation, enabled by inverse visual inference of physical states and latent properties; unified simulation of articulated and deformable objects through its asset-conditioned interface; and sparse-reward robotic manipulation, where differentiable dynamics enable direct analytical policy gradients for efficient policy optimization.



\section{Related Work}

\vspace{8px}
\subsection{Traditional Physics Engines for Robotics}

Simulators such as MuJoCo~\cite{todorov2012mujoco}, PyBullet~\cite{coumans2021}, Gazebo~\cite{gazebo}, robosuite~\cite{robosuite}, and Isaac Gym/Lab~\cite{isaacgym-pre,isaacgym,isaaclab} are the standard for robotic learning due to their high physical fidelity and scalability. 
However, their reliance on non-differentiable numerical solvers introduces discontinuities that prevent gradient propagation through environment transitions.
Differentiable simulators, including Brax~\cite{freeman2021brax}, Warp~\cite{warp2022}, DiffTaichi~\cite{hu2019difftaichi}, and GradSim~\cite{murthy2020gradsim}, seek to address this limitation by redesigning simulation pipelines for automatic differentiation. 
While these methods enable gradient-based optimization, they often trade off physical accuracy, relying on simplified or approximate dynamics. They struggle to faithfully capture complex contact interactions, limiting their applicability in high-fidelity robotic settings.

\subsection{Generative World Models}

Recent generative world models~\cite{agarwal2025cosmos, gao2025adaworld, huang2025vid2world0, rigter2024avid} excel at synthesizing visually coherent futures but often lack explicit physical attributes (\textit{e.g.}, mass, friction) necessary for precise control. 
While state-of-the-art embodied world models such as DreamDojo~\cite{gao2026dreamdojo}, Lingbot-World~\cite{team2026advancing}, MOTUS~\cite{bi2025motusunifiedlatentaction}, and LAWM~\cite{tharwat2025latent} leverage world modeling to improve VLA policies, they do not function as full physical simulators.
Similarly, navigation-centric models~\cite{he2025matrix, ocal2024sceneteller, lu2024infinicube, yu2024wonderworld} focus on infinite scene generation rather than object-centric interactivity.
Fundamentally, these paradigms entangle visual synthesis with temporal transitions, encoding physics implicitly within massive parameter spaces. This entanglement hampers reliable gradient propagation, making them ill-suited for the differentiable robot simulation.

\subsection{Latent World Models}
Latent world models have transformed model-based RL by enabling compact, controllable representations of environment dynamics~\cite{ha2018world}. 
%
Building on this foundation, methods such as PlaNet~\cite{hafner2019planet}, Dreamer~\cite{hafner2025mastering}, and subsequent extensions~\cite{Pan2024IsoDreamPP,li2025open,sun2024learning,burchi2025learning,cohen2024improving} achieve efficient policy optimization by performing imagination and rollout entirely in latent space.
These advances have been further adopted in robotics for manipulation and offline evaluation (\textit{e.g.}, IRASim~\cite{zhu2025irasim}, PIN-WM~\cite{li2025pin}, and WorldGym~\cite{worldgym2025}), often outperforming model-free baselines.
Despite their empirical success, latent world models remain poorly aligned with simulation-oriented settings. They provide limited support for customizable assets, such as variable object geometries or compositional environments. Moreover, their reliance on abstract latent representations deprioritizes high-fidelity visual rendering, restricting their applicability in scenarios that require photorealistic simulation or rendering-based evaluation.

\vspace{8px}
\section{\model{}}

We aim to learn a differentiable world model that functions as an execution engine for embodied manipulation tasks. 
%
%
Unlike monolithic world models that entangle physics and pixels, \model{} adopts a decoupled neural architecture comprising two specialized modules: \textit{\model{}-Dynamics} and \textit{\model{}-Vision}.
As illustrated in~\figref{fig:overview}, this decomposition mirrors the separation of physics and rendering in classical simulation software, and it provides flexible, state-visual dual interfaces for downstream robot learning, supporting both MDP and POMDP formulations.

We first introduce a unified world representation for the robot and scene objects, which accommodates multiple object types and their corresponding geometry and physics descriptors (Sec.~\ref{sec:world_representation}).
At each step, \model{}-Dynamics autoregressively predicts physical states from historical trajectories, actions, and scene descriptors (Sec.~\ref{sec:dynamics}).
\model{}-Vision renders the next observation by grounding latent diffusion in the predicted states and scene context (Sec.~\ref{sec:vision}).
This modularity decouples physical evolution from visual complexity, while maintaining a fully differentiable pipeline that enables real-to-sim system identification and downstream gradient-based policy learning (Sec.~\ref{sec:control_sysid}).

\subsection{Unified Asset and World Representation}
\label{sec:world_representation}

We model embodied manipulation as a physically grounded simulation process in which a robot interacts with $N$ objects over discrete time steps.
Each task instance is defined by a scene configuration specifying object geometry, physical properties, and global environment parameters.
As a drop-in replacement for a classical physics engine, \model{} represents the system through three components:
\begin{itemize}[leftmargin=*]
    \item
    \textit{Physical state ($x_t$):} The time-dependent physical state $x_t$ is defined as the union of the robot's physical state $x_t^p$ and the states of $N$ interacting objects $\{x_t^{(i)}\}_{i=1}^{N}$. Each state vector contains the quantities required for dynamics prediction, such as pose, velocities, joint configurations, or other time-varying state variables depending on the asset type.
    \item
    \textit{Visual observation ($o_t$):} Given $x_t$, the world-model engine produces a high-dimensional visual observation in the form of an RGB image $o_t \in \mathbb{R}^{H \times W \times 3}$.
    \item
    \textit{Static scene descriptors ($\bar{x}$):} We use $\bar{x}$ to represent time-invariant properties that characterize the neural simulator's physical bounds. We categorize these into:
    \begin{itemize}[leftmargin=*]
    \item Asset attributes ($\bar{x}^{(i)}$), which are intrinsic properties of each object $i$, such as geometry, mass, inertia, surface friction, or other asset-specific descriptors.
    \item World configurations ($\bar{x}^w$), which capture global environment parameters shared across all objects, such as gravity or contact properties.
    \end{itemize}
\end{itemize}

A key advantage of this representation is that \model{} is designed as a general-purpose, asset-conditioned framework rather than a task-specific predictor.
For simple objects, physical states and static descriptors can be represented directly using low-dimensional attributes.
For objects with more complex geometry, such as \textit{articulated} or \textit{deformable} assets, their shape descriptions (\textit{e.g.}, meshes or point clouds) can instead be encoded by the corresponding geometric encoder into fixed-shape object-level or part-level tokens, while the overall representation interface remains unchanged.
Additional implementation details for complex objects are provided in Appendix~\ref{app:complex_objects}.


\subsection{\model{}-Dynamics: Object-Centric Physics Core}
\label{sec:dynamics}

\begin{figure}[t]
    \centering
    \includegraphics[width=\linewidth]{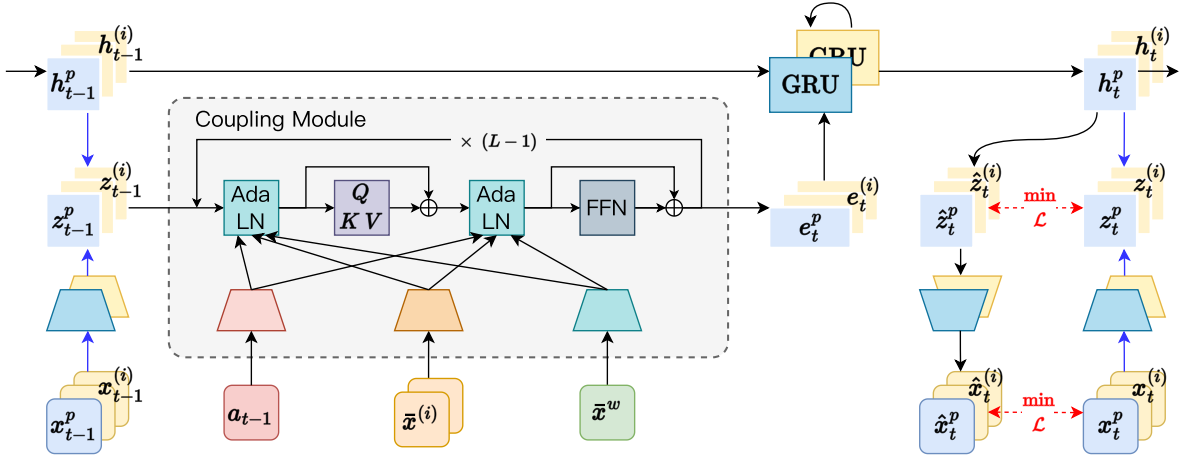}
    \caption{\textbf{\model{}-Dynamics}. An object-centric recurrent dynamics core that predicts next-step physical states. The model utilizes a Transformer-based coupling module to capture multi-entity interactions, with dynamics modulated by control actions and physical attributes via AdaLN.
    }
    \label{fig:dynamics}
\end{figure}

\model{}-Dynamics models the evolution of physical states as a deterministic, history-aware recurrent process.
As shown in~\figref{fig:dynamics}, given the past states $x_{0:t-1}$, the previous action $a_{t-1}$, and static scene descriptors $\bar{x}$, the dynamics core predicts the subsequent state:
\begin{equation} \label{eq:Dynamics}
\hat{x}_t = f_\phi^{\text{dyn}}(x_{0:t-1}, a_{t-1}, \bar{x}).
\end{equation}

To capture the complex multi-object interactions, \model{}-Dynamics employs an \textit{object-centric paradigm}. Each entity (the robot and individual objects) is treated as a discrete token with its own latent and recurrent memory.
Static object attributes, world configurations, and control actions are incorporated as conditioning signals via adaptive normalization, providing structured modulation of the dynamics.
We employ a recurrent state space model structured as follows:
\begin{equation}  
\label{eq:RSSM}
\left\{
\begin{aligned}
& z_t = f^\text{enc}_\phi (h_t, x_t), \quad
h_t = f^\text{rec}_\phi (h_{t-1}, e_t), \quad
\hat{z}_t = f^\text{tra}_\phi (h_t), \quad
\hat{x}_t = f^\text{dec}_\phi (\hat{z}_t), \\
& \textbf{Coupling module:}\quad
e_t = f^\text{cp}_\phi (z_{t-1}, a_{t-1}, \bar{x}).
\end{aligned}
\right.
\end{equation}
The execution begins by mapping the physical state $x_{t-1} = \{ x_{t-1}^p, x_{t-1}^{(1)}, \dots, x_{t-1}^{(N)} \}$ and corresponding recurrent hidden states $h_{t-1} = \{ h_{t-1}^p, h_{t-1}^{(1)}, \dots, h_{t-1}^{(N)} \}$ into a sequence of latent tokens $z_{t-1} = \{ z_{t-1}^p, z_{t-1}^{(1)}, \dots, z_{t-1}^{(N)} \}$, where the first token is for the robot.
These tokens are processed by a \textit{Transformer-based Coupling Module} utilizing global self-attention to model inter-entity dependencies, such as collisions and articulated constraints.
We inject physical constants and control signals through Adaptive Layer Normalization (AdaLN) to modulate the dynamics flow, where the token feature $u$ represents physical states and the condition $c$ encodes actions and static scene descriptors:
\begin{equation} \label{eq:AdaLN}
\mathrm{AdaLN}(u, c) = \gamma(c) \odot \mathrm{LN}(u) + \beta(c),
\end{equation}
where $\gamma(\cdot)$ and $\beta(\cdot)$ are learnable functions of the condition.
Specifically, the robot token is modulated by the action $a_{t-1}$ and world configurations $\bar{x}^w$, while object tokens are modulated by their respective attributes $\bar{x}^{(i)}$ (\eg mass, friction) and $\bar{x}^w$.
%
This structured modulation provides the model with the necessary inductive bias to generalize across varying physical environments.

The coupling module outputs physics-aware embeddings
$e_t = \{ e_t^p, e_t^{(1)}, \dots, e_t^{(N)} \}$, which are then combined with $h_{t-1}$ to update $h_t$.
From the updated recurrent states, a transition module produces the predicted latent states:
$\hat{z}_t = \{ \hat{z}_t^p, \hat{z}_t^{(1)}, \dots, \hat{z}_t^{(N)} \}$,
which are finally decoded into system state predictions:
$\hat{x}_t = \{ \hat{x}_t^p, \hat{x}_t^{(1)}, \dots, \hat{x}_t^{(N)} \}$.

\model{}-Dynamics is trained with three loss terms, including the latent alignment loss ($\mathcal{L}_{\text{tra}}$), the encoding consistency loss ($\mathcal{L}_{\text{enc}}$), and the state reconstruction loss ($\mathcal{L}_{\text{dec}}$):
\begin{equation} \label{eq:3_losses}
    \mathcal{L}_{\text{tra}} = \|\hat{z}_t - \mathrm{sg}(z_t)\|^2 \vphantom{\|}, \quad \mathcal{L}_{\text{enc}} = \|z_t - \mathrm{sg}(\hat{z}_t)\|^2 \vphantom{\|}, \quad \mathcal{L}_{\text{dec}} = \mathcal{L}_{\text{state}}(x_t, \hat{x}_t) \vphantom{\|},
\end{equation}
where $\mathrm{sg}(\cdot)$ denotes gradient stopping.
$\mathcal{L}_{\text{state}}$ may employ different loss forms for different physical state components (see Appendix~\ref{app:dynamics}).
For long-horizon stability, the total objective $\mathcal{L}_{\text{dyn}}$ is optimized over $T$-step autoregressive rollouts over an episode
$\{ \mathcal{D}_t = (x_t, z_t, \hat{x}_t, \hat{z}_t)\}_{t=1}^T$.


\subsection{\model{}-Vision: State-Guided Diffusion}
\label{sec:vision}


\model{}-Vision transforms predicted physical states into high-fidelity visual observations $\hat{o}_t$ through a conditional image generation module.
As shown in Figure~\ref{fig:vision}, We formulate this as a \textit{Latent Diffusion Process}, which is grounded in the scene context $\bar{x}$ and the physical state $\hat{x}_t$ produced by \model{}-Dynamics:
\begin{equation} \label{eq:Vision}
\hat{o}_t \sim p_\phi^{\text{vis}}(o_t \mid o_{t-K:t-1}, \hat{x}_t, \bar{x}),
\end{equation}
where the denoising process is carried out in the latent space and explicitly conditioned on the predicted physical state.
We represent the visual latent at time $t$ and diffusion step $(s)$ as $y_{t,(s)}$, spanning from clean ($s=0$) to noisiest ($s=S$) states.
The preceding $K$ observations are VAE-encoded into clean latents $y_{t-K:t-1,(0)}$, which serve as the context for the next-frame generation. We assemble a sequence $y_{t-K:t}$ by appending an initialization latent to this context, which is set to $y_{t,(0)}$ during training and a zero placeholder during inference.

Once the forward diffusion process reaches the noisy state $y_{t-K:t,(S)}$, the target frame is generated through a denoising process conditioned on the clean context latents.
%
%
First, we achieve \textit{visual anchoring} by concatenating a perturbed version of the most recent latent $y_{t-1,(0)}$ channel-wise to the denoising sequence $y_{t-K:t,(S)}$.
Second, \textit{physical grounding} is incorporated by encoding the predicted physical state $\hat{x}_t$ and static scene descriptors $\bar{x}$ into two complementary representations:
\begin{itemize}[leftmargin=*]
\item \textit{Spatial condition maps:} These provide spatially aligned geometric cues, such as heatmaps or depth-like priors, which are concatenated directly to the U-Net feature maps.
\item \textit{Object-level tokens:} These facilitate cross-attention modulation within the U-Net, enabling the model to attend to specific object states during denoising.
\end{itemize}
The U-Net iteratively refines the sequence for $S$ steps, and the final clean latent $\hat{y}_{t,(0)}$ is decoded into pixel space.
After the denoising process, we take the final clean latent at the last time step as the predicted latent representation $\hat{y}_{t,(0)}$.
Finally, a VAE decoder maps $\hat{y}_{t,(0)}$ back to pixels to produce $\hat{o}_t$, completing a fully differentiable rendering process conditioned on physical states.

We train the conditional denoising network $D_\phi$ as follows, which enforces physical consistency by grounding the denoising process in the predicted states:
\begin{equation} \label{eq:Lvis}
\mathcal{L}_{\text{vis}}=\mathbb{E}_{\sigma,\epsilon}
\Big[
w(\sigma)\,
\big\|
\big(
D_\phi(y_{t-K:t,(0)} + \sigma \epsilon;\, c_t)
\big)_t
-
y_{t,(0)}
\big\|_2^2
\Big],
\end{equation}
where $c_t$ denotes the combined visual and physical conditioning, and
$w(\sigma)$ is a noise-dependent weighting function following standard latent diffusion training.
During training, we sample a noise scale $\sigma$ and Gaussian noise $\epsilon \sim \mathcal{N}(0, I)$, and apply noise to the latent sequence.

\subsection{Gradient-Based Control and System Identification}
\label{sec:control_sysid}

\begin{figure}[t]
    \centering
    \includegraphics[width=\linewidth]{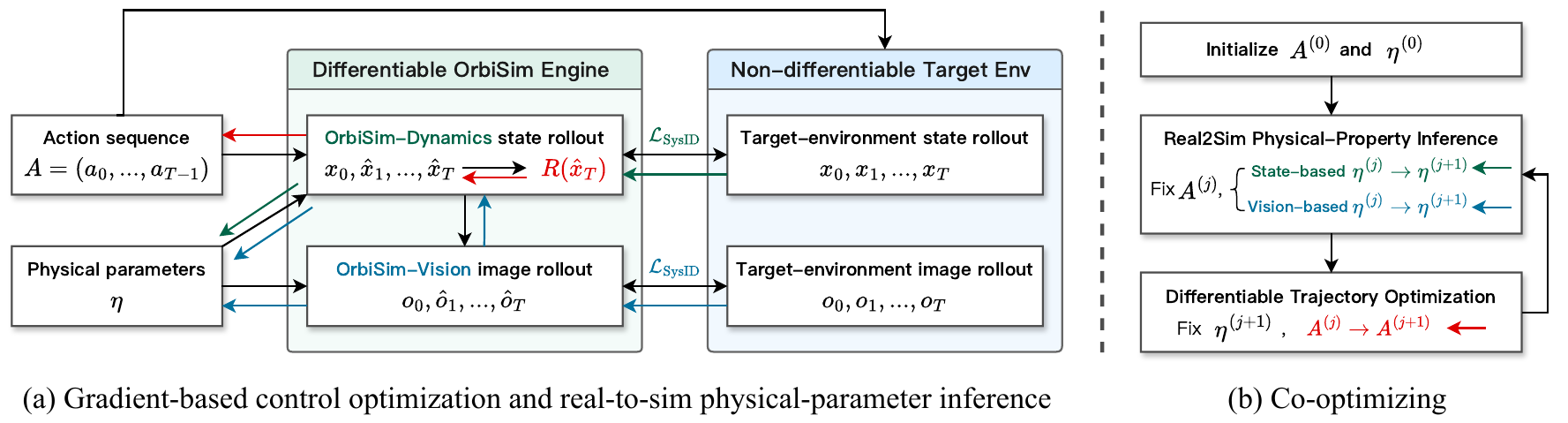}
    \caption{\textbf{Differentiable control and real-to-sim co-optimization.}
    (a) State- or image-space discrepancies between \model{} predictions and corresponding trajectories from the target environment drive physical-parameter inference, while differentiable dynamics provide gradients for action optimization.
    (b) Alternating these updates jointly refines the physical parameters and action sequence.}
    \label{fig:optimization}
\end{figure}

\paragraph{Gradient-Based Control Optimization.}

\begingroup
\setlength{\abovedisplayskip}{5pt plus 1pt minus 1pt}
\setlength{\belowdisplayskip}{5pt plus 1pt minus 1pt}
\setlength{\abovedisplayshortskip}{3pt plus 1pt minus 1pt}
\setlength{\belowdisplayshortskip}{3pt plus 1pt minus 1pt}

The differentiable rollout of \model{} provides action-level sensitivities for optimizing control objectives. 
As shown in Figure~\ref{fig:optimization}(a), let $\mathcal{J}=R(x_T)$ be a terminal reward and let the rollout state evolve according to $\hat{x}_t=f_\phi^{\mathrm{dyn}}(x_{0:t-1},a_{t-1},\bar{x})$. 
The gradient of the reward with respect to each action is
\vspace{4px}
\begin{equation}
\nabla_{a_t}\mathcal{J}
=
\frac{\partial R}{\partial x_T}
\left(\prod_{k=t+1}^{T-1}
\frac{\partial x_{k+1}}{\partial x_k}\right)
\frac{\partial x_{t+1}}{\partial a_t},
\qquad t=0,\ldots,T-1.
\end{equation}

\vspace{4px}
For a differentiable policy $a_t=\pi_\theta(x_t)$, these action gradients can be propagated through the policy parameterization to update its parameters,
\begin{equation}
\theta^{(j+1)}
= 
\theta^{(j)}+\alpha_\theta
\sum_{t=0}^{T-1}
\nabla_{a_t}\mathcal{J}\,
\frac{d a_t}{d\theta},
\end{equation}
\vspace{4px}
or applied directly to an action sequence $A=(a_0,\ldots,a_{T-1})$,
\vspace{4px}
\begin{equation}
A^{(j+1)}
=
\Pi_{\mathcal{A}}\!\left[
A^{(j)}+\alpha_A
\left(\nabla_{a_0}\mathcal{J},\ldots,\nabla_{a_{T-1}}\mathcal{J}\right)
\right].
\end{equation}

\vspace{4px}
Here, $\Pi_{\mathcal{A}}$ enforces the valid action constraints. This formulation allows the same differentiable task objective to support either reusable policy learning or task-specific action refinement, depending on the chosen control parameterization. Because the gradients are computed through the full rollout, a terminal reward supplied only at the end of an episode can still be backpropagated through the differentiable dynamics to all actions in the trajectory.
\endgroup

\vspace{-6pt}
\paragraph{Real-to-Sim Physical-Parameter Inference.}

The integration of \model{}-Dynamics and \model{}-Vision provides a fully differentiable execution engine, enabling gradient-based system identification (SysID) through differentiable rollouts.
Accurate simulation requires scene parameters such as material stiffness, damping, and friction, which are often difficult to measure in real environments. 
Conventional simulators therefore rely on manual calibration or derivative-free search to match simulated and observed trajectories. 
In contrast, \model{} exposes these quantities as optimizable inputs to its differentiable execution graph. 
As shown in Figure~\ref{fig:optimization}(a), let $\eta$ denote the unknown physical parameters contained in the static scene descriptors $\bar{x}$. Given an initial condition and an action sequence $A=(a_0,\ldots,a_{T-1})$, we obtain a reference trajectory from the target environment and a corresponding rollout from \model{} under $\eta$. We identify $\eta$ by minimizing the discrepancy between these trajectories: a state-space discrepancy between reference and predicted state trajectories when physical states are available, or a visual discrepancy between observed and predicted image trajectories when only images are available. The parameter estimate is updated as
\begin{equation}
    \eta^{(j+1)}=\Pi_{\mathcal{E}}\!\left[\eta^{(j)}-\alpha_\eta\nabla_\eta\mathcal{L}_{\mathrm{SysID}}(\eta^{(j)};A)\right],
\end{equation}
where $\mathcal{L}_{\mathrm{SysID}}$ is either the state-space or visual discrepancy and $\Pi_{\mathcal{E}}$ enforces valid parameter ranges. State-based identification provides a privileged reference when states are measurable, whereas vision-based identification uses predicted-observation errors to calibrate parameters that are difficult to measure directly.

\vspace{-6pt}
\paragraph{Co-Optimizing Physical Parameters and Control via Differentiable Rollouts.}
We couple physical-parameter inference and control optimization by alternating between updates to the scene parameters and the action sequence. This allows the learned dynamics to adapt to the observed physical environment while using the adapted model to improve task performance. As shown in Figure~\ref{fig:optimization}(b), starting from an initial action sequence $A^{(0)}=(a^{(0)}_0,\ldots,a^{(0)}_{T-1})$ and a physical-parameter estimate $\eta^{(0)}$, the joint optimization iteratively alternates between the following two steps:

\textit{Real-to-Sim Physical-Parameter Inference.} For a fixed action sequence $A^{(j)}$, we apply the same sequence to the target environment and replay it in \model{}. The discrepancy between the observed and predicted rollouts, measured in state or visual space, is used to update $\eta$ while keeping $A^{(j)}$ fixed.

\textit{Differentiable Trajectory Optimization.} Fix the updated parameters $\eta^{(j+1)}$ and optimize the action sequence with respect to the task objective. Gradients propagated through the \model{}-Dynamics rollout provide the action updates, subject to the valid action constraints.

Repeat the above two steps until convergence or a predefined iteration budget is reached. This co-optimization process enables \model{} to jointly refine the physical parameters and the control trajectory through a unified differentiable rollout framework.

\section{Experiments}

\begin{table*}[b]
\centering
\small
\vspace{-12pt}
\caption{
\textbf{Video-level world modeling performance on the robosuite \texttt{Push} task.}
We report PSNR and LPIPS at different rollout horizons ($10$ / $100$ steps), together with the overall FVD score.
\textit{TrajErr} measures the discrepancy between inferred physical states from generated videos and the corresponding true trajectories.
All learned models perform autoregressive rollouts from shared initial states. The OrbiSim-Dynamics + Oracle Renderer variant is included as a diagnostic reference.
}
\label{tab:wm_push}
\setlength\tabcolsep{3.2pt}
\begin{tabular}{lcccccccc}
\toprule
\textbf{Method}
& \textbf{PSNR10} $\uparrow$
& \textbf{PSNR100} $\uparrow$
& \textbf{LPIPS10} $\downarrow$
& \textbf{LPIPS100} $\downarrow$
& \textbf{FVD} $\downarrow$
& \textbf{TrajErr} $\downarrow$\\
\midrule
Vid2World
&22.2014  &17.8856  &0.1312  &0.2551  &1750.1 & 0.6754  \\

AdaWorld
&26.6647  &12.8346  &0.1183  &0.3482  &1305.8 & 1.8597  \\
\midrule
Orbisim w/o Decoupling
&\textbf{27.9346}  &\underline{19.9510}  &0.1188  &0.1799  &689.1 & 0.8134 \\

Orbisim w/o Random Sampling
&26.6890  &19.1119  &\textbf{0.1076}  &0.1669  &\underline{531.2} & 0.5742 \\

Orbisim w/o Object-Centric
&25.9373  &19.7581  &0.1123  &\underline{0.1463}  &\textbf{524.5} & \underline{0.4687} \\

\textbf{Orbisim (Final)}
&\underline{26.7105}  &\textbf{19.9819}
&\underline{0.1078}  &\textbf{0.1428}  & 533.9 & \textbf{0.4468} \\

\midrule

\makecell[l]{OrbiSim-Dynamics +\\ Oracle Renderer}
&30.7160
&27.8247
&0.0163
&0.0279
&39.1236
&0.1133 \\

\bottomrule
\end{tabular}
\end{table*}

We evaluate \model{} as both a generative world model and a differentiable execution engine, focusing on (1) generative fidelity and physical consistency under varying configurations, and (2) the benefits of differentiable gradient pathways for downstream policy learning and SysID.


\subsection{Experimental Configurations and Baselines}

We evaluate \model{} on five physics-rich environments built upon classical simulation platforms:
$\bullet$ \textbf{Robosuite \texttt{Push}:} A UR5e manipulator pushes one cube into another under randomized object and physical configurations~\citep{robosuite}.
$\bullet$ \textbf{Isaac Lab \texttt{Stack}:} A Franka arm sequentially stacks three cubes, testing long-horizon stability and multi-object interaction~\citep{isaaclab}.
$\bullet$ \textbf{AdaManip \texttt{Articulated}:} A robot manipulates articulated objects with diverse shapes and joint mechanisms in the AdaManip environment~\citep{Wang2025AdaManipAA,isaacgym}.
$\bullet$ \textbf{Physion \texttt{Drape}:} A cloth interacts with rigid objects of varying geometries in the Physion Drape scenario~\citep{Bear2021PhysionEP}.
$\bullet$ \textbf{MuJoCo \texttt{Cable Lift}:} Two robot arms coordinate to move designated points on a deformable cable toward target locations, with cable stiffness as a key physical factor~\citep{todorov2012mujoco}.
More dataset details are provided in Appendix~\ref{app:data_collection}.

We organize our evaluation into three groups. First, for world-model generation quality (Sec.~\ref{sec:wm_eval}), we compare \model{} with Vid2World~\citep{huang2025vid2world0} and AdaWorld~\citep{gao2025adaworld}. Second, for downstream policy performance (Sec.~\ref{sec:rl_eval}), we compare with representative model-free and model-based RL methods, as well as behavior cloning; detailed experimental settings are provided in Appendix~\ref{app:rl}. Finally, to demonstrate the advantage of differentiability (Sec.~\ref{sec:physical_property_policy}), we evaluate the MuJoCo \texttt{Cable Lift} task in both simulation and the real world. We compare against $\pi_{0.5}$ and SAC in simulation, and deploy $\pi_{0.5}$ as the baseline on an AgileX dual-arm platform.

\subsection{Evaluating Generative and Physical Fidelity}
\label{sec:wm_eval}




\subsubsection{Performance on Benchmark Manipulation Tasks}
\vspace{-2pt}

\begin{figure*}[!t]
    \centering
    \begin{minipage}{0.495\textwidth}
        \centering
        \includegraphics[width=\textwidth]{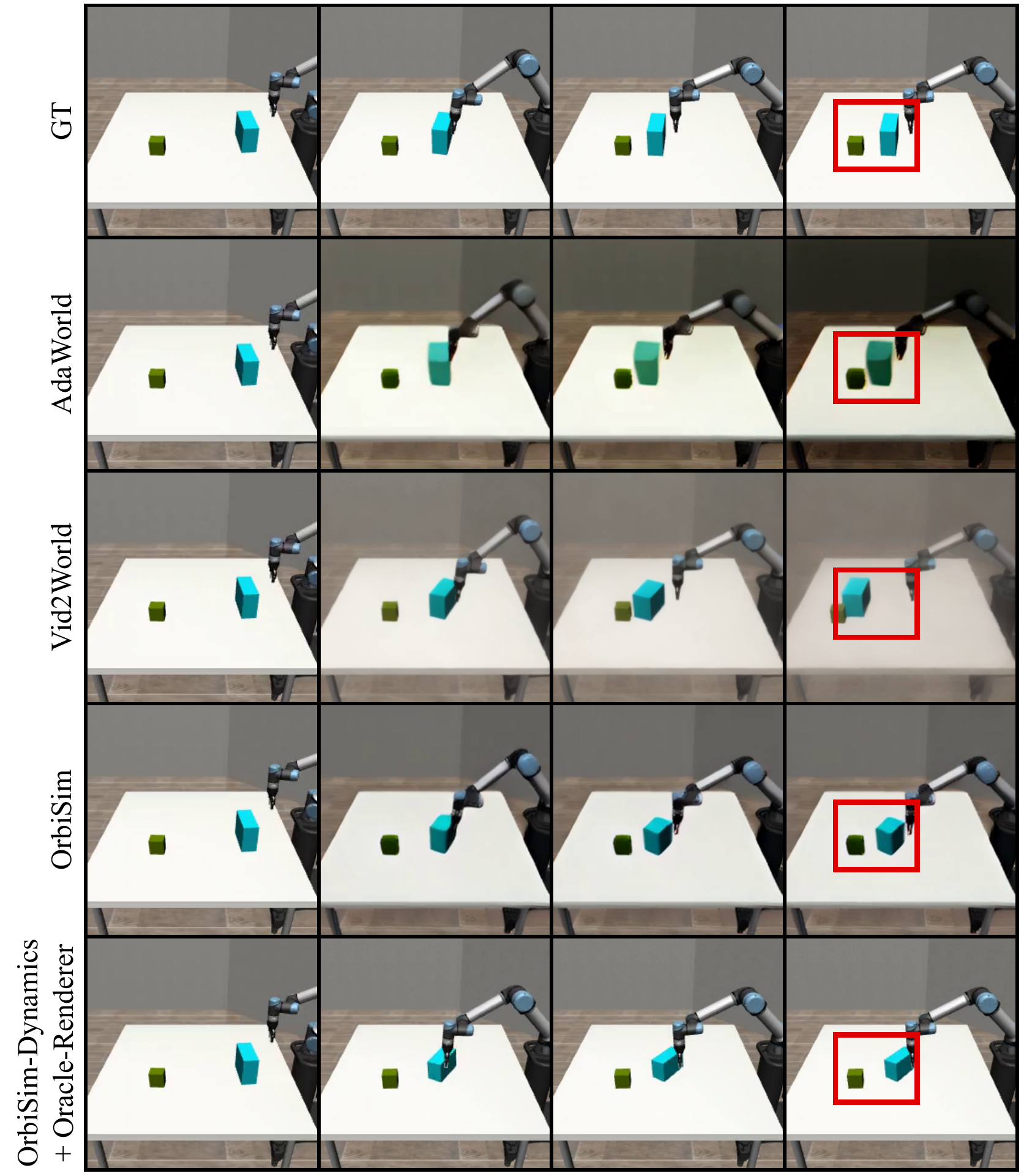}
        \par\small (a) High friction
    \end{minipage}
    \hfill
    \begin{minipage}{0.495\textwidth}
        \centering
        \includegraphics[width=\textwidth]{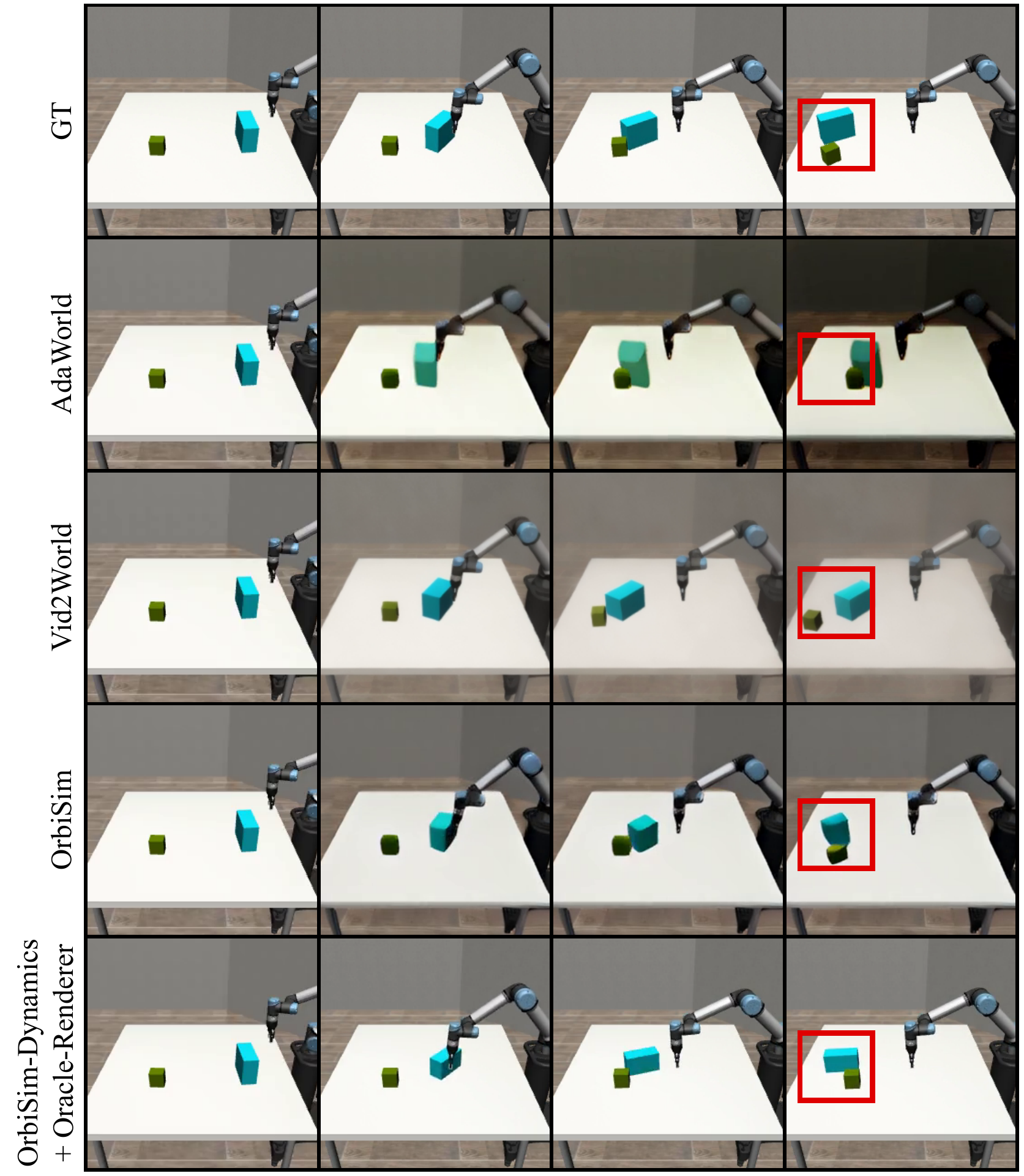}
        \par\small (b) Low friction
    \end{minipage}
    
    \vspace{-6pt}\caption{\textbf{Autoregressive simulation under varying physical parameters.}
    All methods use the same initial observations and action sequences.
    The ``OrbiSim-Dynamics + Oracle-Renderer'' variant renders the physical states predicted by \model{}-Dynamics with the underlying simulator renderer, replacing \model{}-Vision to isolate state-transition prediction from visual-rendering errors.
    The comparison shows that \model{} better captures system dynamics under different object-table friction settings.}
    \label{fig:vis_friction}
    \vspace{-12pt}
\end{figure*}

\begin{figure*}[t]
    \centering
    \includegraphics[width=\textwidth]{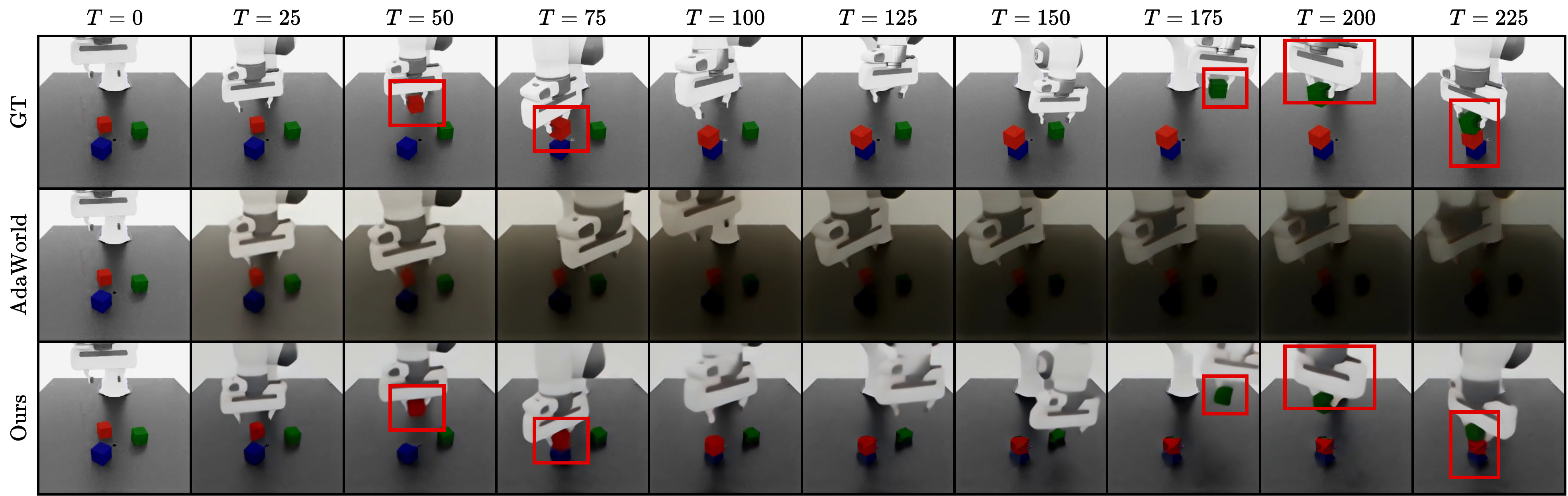}
    \caption{\textbf{Autoregressive simulation on the Isaac Lab \texttt{Stack} task over a $225$-step horizon.}
    \model{} accurately captures intricate manipulator rotations and complex multi-object collisions, maintaining high physical fidelity and stability throughout the sequence.
    }
    \label{fig:vis_stack}
\end{figure*}

As shown in \tabref{tab:wm_push}, \model{} (Final) consistently achieves state-of-the-art performance across all metrics and horizons. We compare against \textbf{AdaWorld}~\citep{gao2025adaworld} and \textbf{Vid2World}~\citep{huang2025vid2world0}; while these baselines produce visually plausible frames, they struggle with compounding physical drift. In contrast, \model{} maintains superior temporal coherence (FVD) and lower trajectory error, demonstrating a more robust alignment between physical dynamics and visual synthesis.
To isolate the quality of state-transition modeling from learned visual rendering, we additionally report an \textit{OrbiSim-Dynamics + Oracle Renderer} variant. This variant renders the physical states predicted by \model{}-Dynamics with the underlying simulator renderer, providing a diagnostic evaluation of the predicted dynamics. Because it uses an oracle renderer, its image-level metrics are not directly comparable to those of the learned visual models; instead, the result measures how accurately the predicted state trajectories support physically consistent rendering.
Qualitative visualizations in \figref{fig:vis_friction} and \figref{fig:vis_stack} further suggest that \model{} remains robust across varying physics coefficients and complex stacking dynamics over a $225$-step long-term simulation horizon.

\vspace{-4pt}
\subsubsection{Ablation Studies}
\vspace{-2pt}

We assess the contribution of core design choices through three variants:
$\bullet$
\emph{\model{} w/o Decoupling} removes the separation between dynamics and vision, training a single monolithic model to jointly predict states and observations.
$\bullet$
\emph{\model{} w/o Random Sampling} disables random sparsification of visual context during training, using only contiguous context frames.
$\bullet$
\emph{\model{} w/o Object-Centric} replaces object-centric dynamics modeling with a global state representation that concatenates the robot state, all object states, object attributes, and world configurations at each time step.

As shown in Table~\ref{tab:wm_push}, removing the decoupled dynamics–vision architecture improves single-frame reconstruction (higher PSNR) but severely degrades temporal consistency and trajectory accuracy, leading to pronounced long-horizon distortions, confirming the necessity of separating dynamics modeling from visual rendering. Disabling random visual context sampling slightly degrades most metrics and increases trajectory error, indicating that context sparsification improves robustness to autoregressive noise. Replacing object-centric modeling with a global state representation yields consistent but moderate performance drops, with visualizations further revealing impaired modeling of object interactions, highlighting the importance of object-level structure for multi-body dynamics.

\vspace{-4pt}
\subsubsection{Generalization to Heterogeneous Assets}
\vspace{-2pt}

As discussed in Sec.~\ref{sec:world_representation}, \model{} is designed as a general-purpose, asset-conditioned framework that can accommodate heterogeneous object types through appropriate state and asset encodings.
We therefore evaluate this extensibility on two additional complex-object settings: articulated-object manipulation from AdaManip~\citep{Wang2025AdaManipAA} and deformable cloth dynamics from the Physion Drape scenario~\citep{Bear2021PhysionEP}. 
For articulated objects, object assets are represented by structured 3D asset files, such as URDF and mesh files, together with joint states. 
For deformable objects, cloth geometry is represented using point-based states. 
We encode these geometry-conditioned descriptors with a geometric encoder, such as Point Transformer V3~\citep{Wu2023PointTV}, and feed the resulting asset tokens into the same object-centric dynamics and state-guided vision modules. 
More implementation details are provided in Appendix~\ref{app:complex_objects}.

Figure~\ref{fig:complex_objects} shows qualitative autoregressive rollouts on both settings. 
\model{} accurately follows joint-constrained part motion in articulated-object manipulation and captures the overall deformation pattern of cloth interacting with rigid objects. 
These results suggest that the asset-conditioned design enables \model{} to accommodate substantially richer object geometries and physical properties without redesigning the simulator architecture.

\begin{figure*}[!t]
    \centering
    \begin{minipage}{0.495\textwidth}
        \centering
        \includegraphics[width=\textwidth]{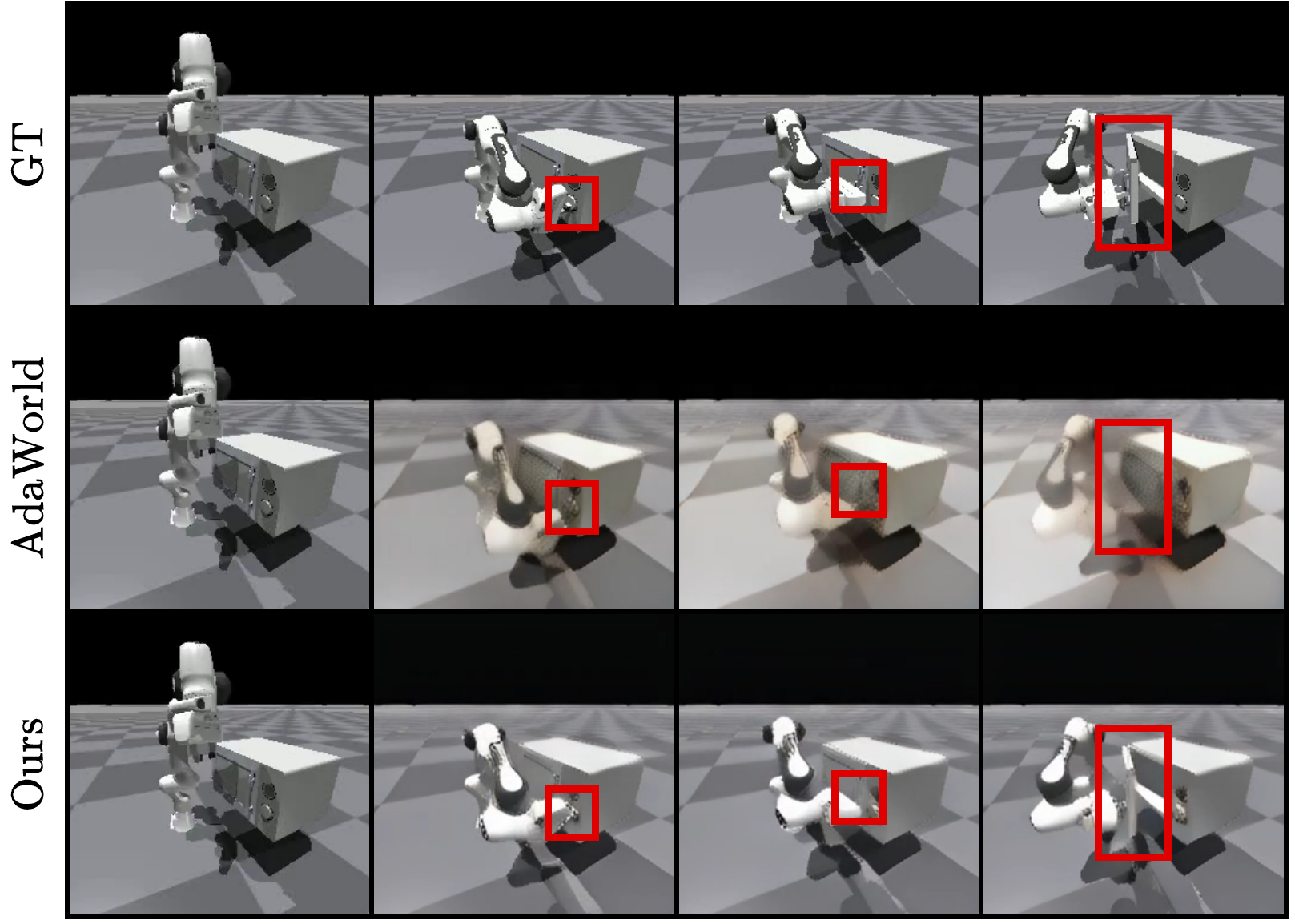}
        \par\small (a) AdaManip \texttt{Articulated}
    \end{minipage}
    \hfill
    \begin{minipage}{0.495\textwidth}
        \centering
        \includegraphics[width=\textwidth]{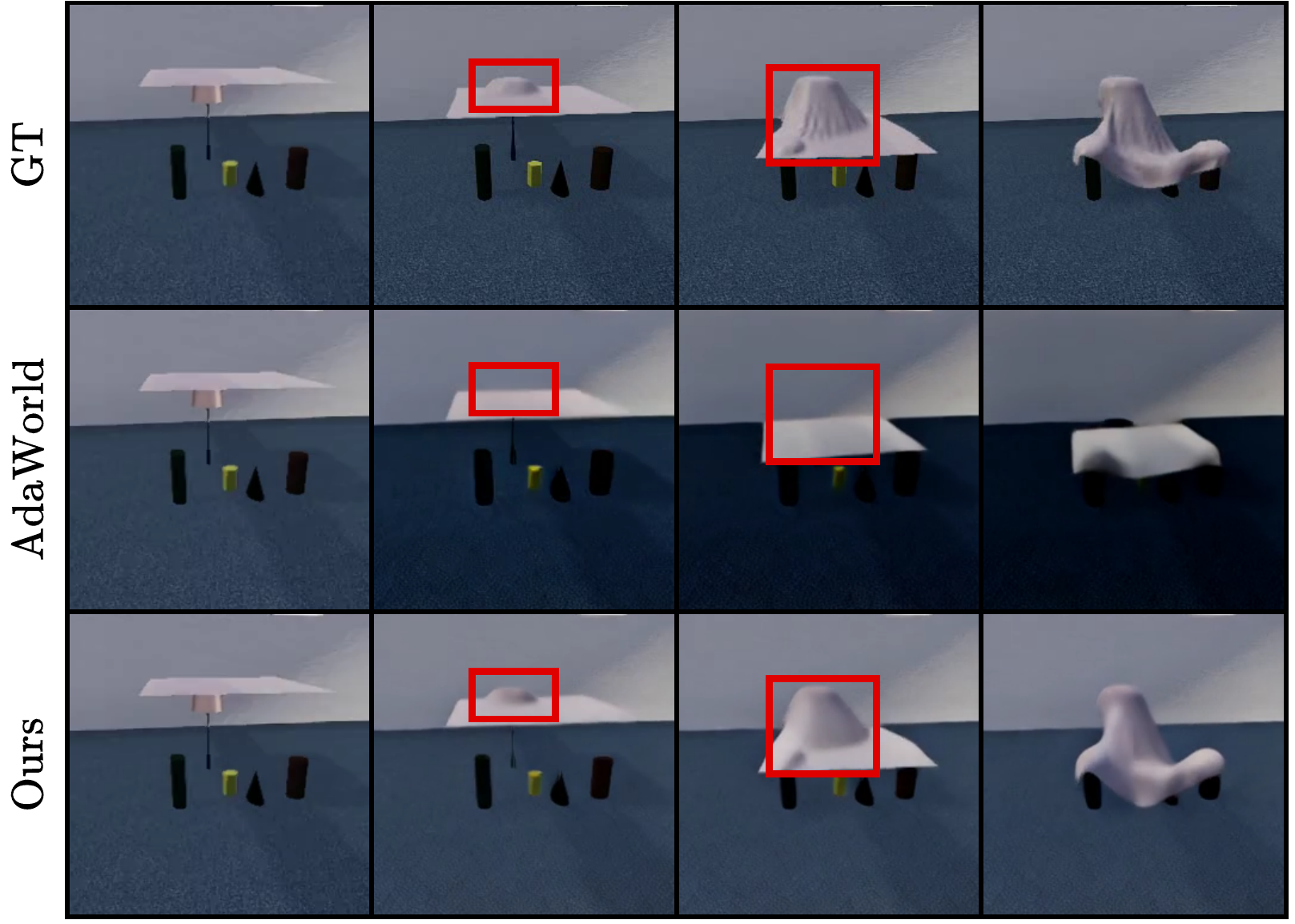}
        \par\small (b) Physion \texttt{Drape}
    \end{minipage}
    \caption{\textbf{Autoregressive simulation on complex objects.} (a) Articulated-object manipulation from AdaManip.  (b) Deformable cloth dynamics from the Physion Drape scenario.  \model{} captures both joint-constrained articulated motion and geometry-conditioned cloth deformation under the same asset-conditioned simulation framework.}
    \label{fig:complex_objects}
    \vspace{-8pt}
\end{figure*}

\begin{table*}[!t]
\centering
\small
\caption{
\textbf{Out-of-distribution world modeling performance.}
We report PSNR and LPIPS at different rollout horizons ($10$ / $100$ steps), together with the overall FVD score.
OOD settings include unseen object counts, unseen shape distributions, and extreme physical parameters.
}
\label{tab:ood_generalization}
\setlength\tabcolsep{8pt}
\begin{tabular}{lccccc}
\toprule
\textbf{Method}
& \textbf{PSNR10} $\uparrow$
& \textbf{PSNR100} $\uparrow$
& \textbf{LPIPS10} $\downarrow$
& \textbf{LPIPS100} $\downarrow$
& \textbf{FVD} $\downarrow$\\
\midrule
AdaWorld (ID)
&26.6647  &12.8346  &\underline{0.1183}  &0.3482  &1305.8 \\

\model{} (ID)
&\underline{26.7105}  &\underline{19.9819}
&\textbf{0.1078}  &\textbf{0.1428}  &\textbf{533.9} \\

\midrule
AdaWorld (OOD)
&26.3468  &13.2215  &0.1228  &0.3836  &1288.5 \\

\textbf{\model{} (OOD)}
&\textbf{27.1867}  &\textbf{20.1109}
&0.1188  &\underline{0.1847}  &\underline{597.3} \\
\bottomrule
\end{tabular}
\end{table*}

\subsubsection{Robustness to Out-of-Distribution Scenarios}

We further evaluate the generalization ability of \model{} under out-of-distribution (OOD) configurations. We construct three types of OOD scenarios: $\bullet$ \textit{Object-count OOD}, where the model is trained on two-object scenes and evaluated on unseen three-object scenes; $\bullet$ \textit{Shape OOD}, where point-cloud-based object geometries in the test split do not appear during training; and $\bullet$ \textit{Physical-parameter OOD}, where we evaluate the model under extreme friction settings that are significantly outside the training distribution in the push-hit task.

Table~\ref{tab:ood_generalization} summarizes the video-level performance under in-distribution and OOD settings. Even under unseen object counts, shapes, and physical conditions, \model{} maintains strong generation quality and substantially outperforms AdaWorld, especially over long horizons. These results indicate that the asset-conditioned and object-centric design of \model{} provides robust generalization beyond the training distribution, rather than merely memorizing seen scene configurations.

\subsection{Real-to-Sim Physical Inference and Real-Robot Action Optimization}
\label{sec:physical_property_policy}

Manipulating deformable objects such as cables in the real world poses a challenge for robotic control. Cable motion involves high-dimensional, nonlinear deformation that depends on unknown material properties, such as bending stiffness, damping, and elastic recovery. Dense, frame-by-frame measurements of the cable's physical state are also difficult to obtain outside simulation.

\begin{figure*}[t]
    \centering
    \includegraphics[width=0.9\textwidth]{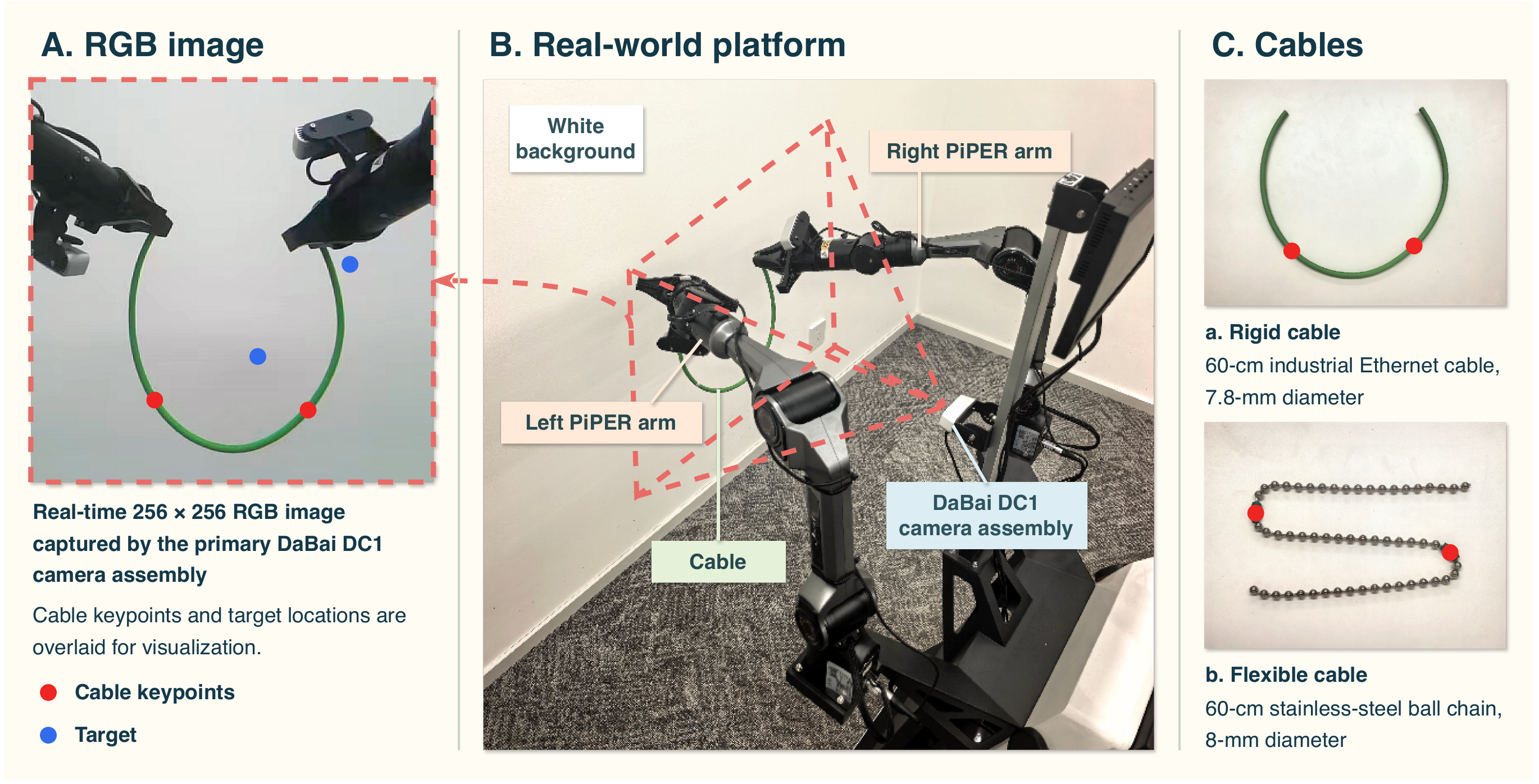}
    \caption{\textbf{Real-robot experimental setup.}
    (A) A real-time RGB image captured by the primary DaBai DC1 camera assemblyn.
    (B) The bimanual PiPER platform, camera assembly, and cable arrangement.
    (C) Two cables with different material properties. The cable keypoints are positioned at one-third and two-thirds of the cable length.}
    \label{fig:real-world-setting}
\end{figure*}

As shown in Figure~\ref{fig:real-world-setting}, we study this challenge through a bimanual \texttt{Cable Lift} task on a real AgileX robot platform. The two arms coordinate to move designated cable keypoints toward their target locations. Leveraging the end-to-end differentiability of \model{}, we connect physical-parameter inference from real-robot RGB observations and recorded actions to downstream action-trajectory optimization, without requiring ground-truth cable states. We first evaluate the resulting actions on the real robot and then use controlled simulation experiments to examine the inference and optimization procedures in greater detail.

\vspace{-4pt}
\subsubsection{Real-Robot Evaluation}
\vspace{-2pt}

\begin{figure*}[!t]
    \centering
    \includegraphics[width=\textwidth]{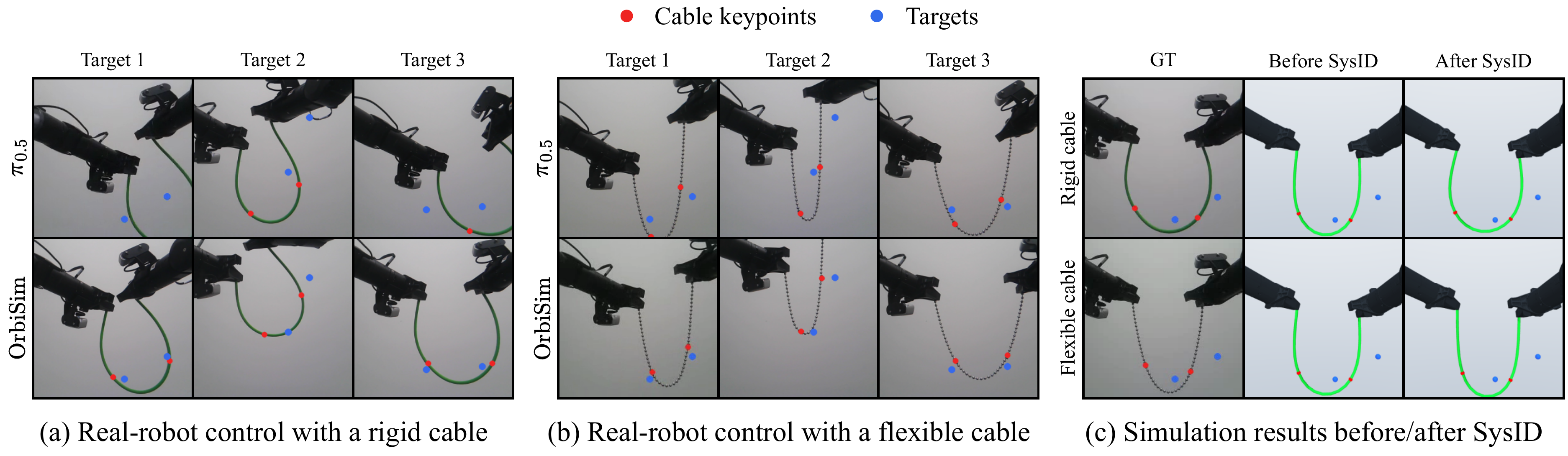}
    \caption{\textbf{Real-world evaluation results on the \texttt{Cable Lift} task.}
    We compare the deployed $\pi_{0.5}$ policy with action sequences optimized using \model{} on two cables with different material properties under three target configurations.
    We also show comparisons between real-world observations (GT) and the corresponding \model{} predictions before and after real-world system identification (SysID) for rigid and flexible cables.
    The optimized actions achieve smaller final keypoint-to-target distances than $\pi_{0.5}$, while SysID improves the agreement between predicted and observed cable configurations.}
    \label{fig:real-cable}
\end{figure*}

\begin{figure*}[t]
    \centering
    \includegraphics[width=0.9\textwidth]{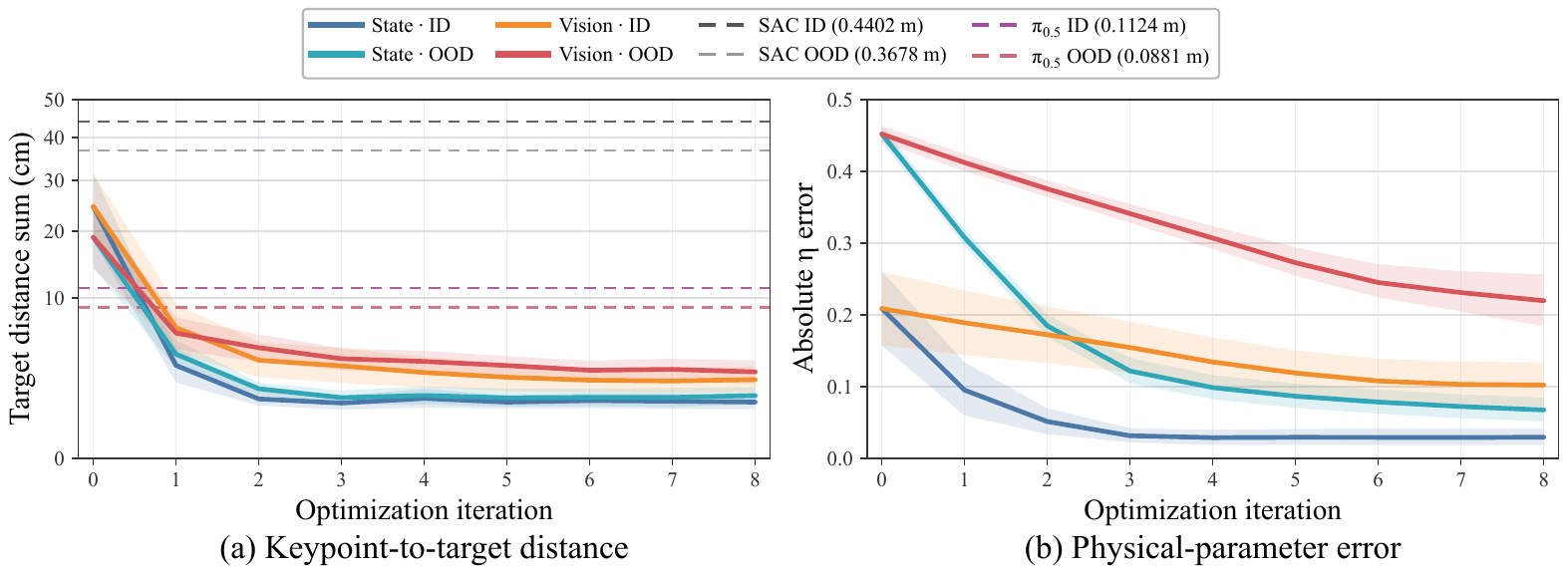}
    \caption{\textbf{Simulation results on the \texttt{Cable Lift} task.} Terminal keypoint-to-target distance and physical-parameter error over eight alternating optimization iterations. State/Vision denote state- and image-based SysID on the ID/OOD splits.}
    \label{fig:cable_simulation_results}
\vspace{-4pt}
\end{figure*}

We evaluate the \texttt{Cable Lift} task on an AgileX dual-arm platform with two cables of different material properties. For each cable, we first collect RGB trajectories together with the executed robot motions and use them to estimate the physical parameters $\eta$. We then fix $\eta$, optimize a fixed-length action sequence through the differentiable \model{}-Dynamics rollout with the terminal keypoint-distance objective, and execute the optimized sequence on the robot. We compare this pipeline with the deployed $\pi_{0.5}$ policy.

Figure~\ref{fig:real-cable} shows the real-world task executions for both control methods under different target configurations, together with comparisons between real observations and \model{} predictions before and after SysID. The optimized actions bring the cable keypoints closer to their targets than $\pi_{0.5}$ on both cables.
The \model{}-optimized actions achieve a mean final keypoint-to-target distance of \textbf{$\mathbf{17.5 \pm 6.1}$ pixels}, compared with \textbf{$\mathbf{62.2 \pm 36.3}$ pixels} for $\pi_{0.5}$, while real-world SysID improves the agreement between predicted and observed cable configurations. By differentiably optimizing the physical parameters and the action sequence, \model{} adapts its learned dynamics to cables with different material properties and improves control without direct cable-state measurement.

\subsubsection{Controlled Simulation Diagnostics}

To further examine the convergence behavior of the alternating inference-and-optimization procedure under controlled conditions, we conduct systematic diagnostics on the \texttt{Cable Lift} task in simulation. Unlike the real-world experiments, the simulator provides ground-truth physical parameters and state trajectories, allowing us to compare state-based and vision-based supervision under matched conditions.

We evaluate the procedure separately on in-distribution (ID) and out-of-distribution (OOD) settings. ID settings use cable properties within the physical-parameter distribution observed during training, whereas OOD settings use properties outside that distribution. For both settings, the state-based and vision-based variants use the same initial configurations and targets. The state-based variant identifies the physical parameters from reference state trajectories, while the vision-based variant uses image-trajectory discrepancies for identification; both variants then optimize the action sequence with the terminal keypoint-distance objective through the differentiable \model{}-Dynamics rollout.

Figure~\ref{fig:cable_simulation_results} summarizes the optimization over eight alternating optimization iterations introduced in Sec.~\ref{sec:control_sysid}. The state-based variant provides a privileged reference and rapidly reduces both errors in the ID and OOD settings. The vision-based variant also improves consistently across both splits, although its parameter error remains higher than that of the state-based variant, particularly in the OOD setting. Its final terminal distance reaches \textbf{0.024 m} on ID and \textbf{0.029 m} on OOD, compared with \textbf{0.012 m} and \textbf{0.015 m} for the state-based variant, respectively.
The dashed lines in the distance panel show the corresponding SAC and $\pi_{0.5}$ reference performances. Both differentiable optimization variants reduce the terminal distance below these references during the alternating procedure. These results show that image-trajectory discrepancies can be backpropagated end-to-end through \model{}-Vision and \model{}-Dynamics to provide useful gradients for adapting the physical parameters. The resulting estimates are sufficient to improve downstream action optimization even when the physical properties are outside the training distribution.

Further analyses of the \texttt{Cable Lift} task, including parameter-identification errors and final action-execution distances, are provided in Appendix~\ref{app:cable}.

\subsection{Sparse-Reward Policy Optimization via Differentiable Dynamics}
\label{sec:rl_eval}

\begin{wrapfigure}{r}{0.4\textwidth} 
    \centering
    \small
    \vspace{-10pt}
    \begin{minipage}[c]{0.4\textwidth}
        \centering
        \includegraphics[width=\textwidth]{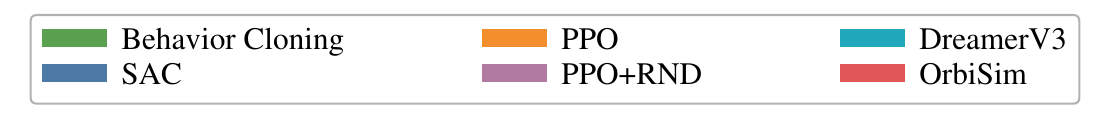}
    \end{minipage}
    \begin{minipage}[c]{0.4\textwidth}
        \centering
        \includegraphics[width=\textwidth]{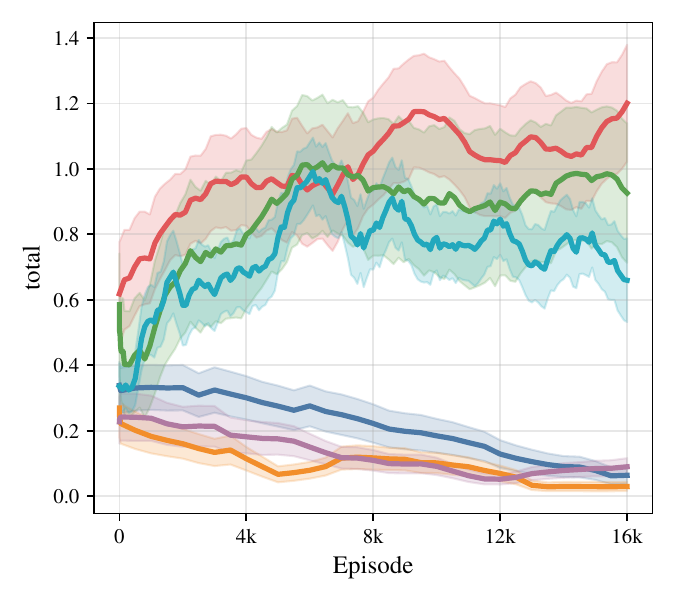}
    \end{minipage}
    \vspace{-8pt}
    \caption{\textbf{Reward curve of robosuite \texttt{Push} task RL.} We compared \model{} with model-free baselines (SAC, PPO, PPO+RND), Behavior Cloning as well as model-based baseline (DreamerV3).}
    \label{fig:rl_training_curves}
    \vspace{-15pt}
\end{wrapfigure}
We evaluate \model{} as a functional execution engine for downstream RL tasks. 
$\bullet$
In the robosuite \texttt{Push} task, the robot must push one cube to hit another into a target region without either object falling off the table. 
A rollout is considered successful if the second cube moves more than $5$ cm.
We define three normalized rewards based on distance reduction to encourage the robot to approach and push the cubes (see Appendix~\ref{app:rl}). 
To preserve sparse supervision, these rewards are provided only at the end of each episode during training. During evaluation, they are used to measure task completion performance.
$\bullet$
In the Isaac Lab \texttt{Stack} task, the agent must sequentially stack three cubes, with binary success determined by whether the final stacked configuration is achieved.

While the sparse episodic reward design makes credit assignment particularly challenging, \model{} differs from traditional black-box simulators by exposing analytical gradient pathways that propagate task-specific reward signals directly to the policy parameters.
As shown in \figref{fig:rl_training_curves}, \model{} achieves superior performance and convergence speed compared to model-free and imitation baselines.
\begin{itemize}[leftmargin=*]
\item \textit{vs. Model-free RL}: We adopt SAC, PPO, and PPO with RND as model-free RL baselines. As shown in the results, model-free methods struggle under extremely sparse rewards, often failing to obtain consistent learning signals and collapsing to near-zero performance. In contrast, \model{} provides informative optimization signals through its differentiable dynamics, enabling the policy to reason about the physical consequences of its actions even under terminal-only supervision.
\item \textit{vs. Behavior cloning}: While BC learns the initial pushing behavior from expert demonstrations, it struggles to generalize to OOD states throughout long-horizon interactions. In contrast, \model{} enables the agent to refine its policy through rollouts in a physical world model.
\item \textit{vs. Model-based RL}: Model-based RL methods such as DreamerV3 predict future dynamics and optimize policies in latent space. However, DreamerV3 lacks explicit physical modeling, which limits its robustness under varying physical settings.
\end{itemize}
Detailed training reward curves and the success rate table are listed in Appendix~\ref{app:rl}.

\subsection{Additional Evaluation}

In Appendix~\ref{app:more_exprs}, we provide comprehensive supplementary analyses across three primary axes:
\begin{itemize}[leftmargin=*]
    \item
    \textit{Component analyses:} We investigate the isolated predictive performance of \model{}-Dynamics (\ref{app:dynamics_eval}) and assess various inference workflows (\ref{app:worlflow}).
    \item
    \textit{Scalability/robustness}: We evaluate \model{}’s scalability to multi-object/multi-task settings (\ref{app:multi}) and its robustness under noisy training data that simulates real-world acquisition pipelines (\ref{app:robustness}).
    \item
    \textit{Extended policy studies:} We offer further insights into policy optimization, including imitation learning baselines (\ref{app:imitation_learning}), comparisons against specialized differentiable simulators such as NVIDIA Newton (\ref{app:newton}), expanded RL experiments (\ref{app:rl}), supplementary simulation and real-world results on the \texttt{Cable Lift} task (\ref{app:cable}), and computational efficiency and resource usage analyses (\ref{app:compute}).
\end{itemize}




\section{Conclusions, Limitations, and Future Work}
\label{sec:conclusion}

In this paper, we proposed OrbiSim, a novel world model paradigm that rethinks simulation from a learning-centric perspective by treating the world model itself as a fully differentiable physics engine.
Through extensive experiments on robotic simulation and control tasks, we demonstrated that OrbiSim outperforms existing world models in future prediction accuracy and control performance.

Despite these encouraging results, OrbiSim still has important limitations. Although it shows promising robustness within related simulator families and already supports heterogeneous assets, its generalization to broader robot morphologies, scene sources, and real-world domains remains underexplored, and it is still far from a complete, general-purpose industrial simulator. Nonetheless, its fully differentiable formulation and scalability in data-rich settings make OrbiSim a promising step toward practical deployment.

Looking ahead, we aim to improve OrbiSim's ability to generalize across physical environments and asset families. Advanced multimodal foundation models could help parse real scenes and extract structured geometric, kinematic, and physical information for constructing training and evaluation environments. A unified tokenizer for common 3D asset formats, including XML, USD, and URDF, could further expand the range of assets that OrbiSim can represent. More broadly, combining the semantic priors of large models with the physically grounded, differentiable rollouts of \\model{}-Dynamics may provide a scalable route to modeling previously unseen scenes and interactions. Together, these directions point toward a sim-to-sim-to-real framework in which knowledge transfers across simulators before being adapted to real-world control.

\bibliographystyle{plain}
\bibliography{ref}

\newpage
\appendix
\section*{Appendix}

\section{Data Collection and Generation}
\label{app:data_collection}

In this appendix, we provide detailed descriptions of the data collection and generation procedures used in our experiments.
These details complement the high-level experimental setups presented in the main paper and are included here for completeness and reproducibility.

\paragraph{Data Representation.} 
Across all environments, datasets are organized as episodic trajectories consisting of structured state observations, static environment configurations, and, when applicable, control actions. 
For robot manipulation tasks, actions are represented as continuous end-effector commands. Each action is a 7-dimensional vector containing translational and rotational commands together with a gripper control signal. 
The Physion \texttt{Drape} task, however, involves passive deformable dynamics of a falling cloth and does not include actions. 
State observations adopt an object-centric representation that factorizes the world state into robot, rigid-object, articulated-object, or deformable-object components, depending on the task. 
Static configurations encode asset descriptors and physical attributes, including geometry, joint structure, friction, density, and global scene parameters. 
This unified episodic format allows \model{} to handle both action-conditioned manipulation and passive physical prediction within the same data pipeline.

\subsection{Robosuite \texttt{Push} Task}

Based on the robosuite simulation engine, we construct a trajectory dataset for a cube pushing and hitting task.
In each simulation trial, several physical parameters were randomized, including cube friction, density, size, and visual appearance. The pushing velocity of the manipulator was also varied, as it significantly affects contact dynamics and task outcomes.

The task involves two cubes placed on a table and a UR5e manipulator. During an episode, the manipulator pushes the first cube, which then collides with the second cube.
A rollout is considered successful if the first cube remains on the table and the second cube reaches a designated region near the target edge without falling off the other edges.

To ensure broad coverage of the task domain, the dataset was collected using a fixed multi-stage policy that generates trajectories from three distinct phases:
(a) random manipulator movements,
(b) pushing a cube toward a random direction,
and (c) pushing a cube to hit another cube.
Specifically, the manipulator first moves to a position behind the first cube, then executes a push toward a sampled direction, and finally remains stationary.

During the simulation, we recorded the full physical states of both cubes and the manipulator, the executed control actions, as well as RGB images captured from a side-view camera.
In total, the dataset contains 120 trajectories from phase (a), 640 trajectories from phase (b), and 1,840 trajectories from phase (c). All trajectories feature randomized physical properties and visual appearances to promote diversity and robustness.

\subsection{Robosuite Tasks with Multi-Objects}
\label{app:robosuite_multi}
To introduce scalability and variety to the robosuite \texttt{Push} task, we subsequently construct two more tasks in robosuite. 

The first task, \texttt{Push\_cubes}, is an extension of the robosuite \texttt{Push} task, which also involves pushing and hitting cubes. The task is separated into four subtasks, namely, (a) randomly pushing (random), (b) pushing stacked cubes (stacked), (c) pushing toward an object (towards\_object), and (d) pushing to hit an object (to\_hit). The first two subtasks have randomized 3 to 4 objects, extending the number of objects. After finishing the task, the robot arm is set to randomly move with smoothed trajectories in order to extend the robot arm position coverage. 

The second task is robosuite \texttt{Pick\_and\_place}, which requires the gripper to randomly pick cubes and place them. This task is separated into three subtasks: (a) stack cubes, (b) unstack initially stacked cubes, and (c) pick and place scattered cubes. The cubes' positions, robot arm picking order, moving keypoint trajectories, and cubes dropping heights are randomized during rollout. 

The above two tasks contain most of the interaction types of robot arms in robosuite, including pushing, hitting, falling, gripper picking, etc. Each subtask of the two tasks has 420 episodes collected.

\subsection{Isaac Lab \texttt{Stack} Task}

We additionally construct a trajectory dataset based on the Isaac Lab cube stacking task, where a Franka manipulator is required to sequentially stack multiple cubes in a physics-based simulation environment (\texttt{Isaac-Stack-Cube-Franka-IK-Rel}).

The dataset consists of episodic trajectories collected from interactions in the simulation, with an average episode length of approximately 230 timesteps. Each trajectory records the full sequence of states and actions throughout an episode.
The final dataset includes approximately 1,000 successful trajectories and 1,200 failure trajectories, providing balanced coverage of diverse task execution outcomes.

The dataset is derived from a small set of human expert demonstrations.
Starting from these demonstrations, we follow the standard Isaac Lab Mimic pipeline to automatically generate additional state-based trajectories, substantially increasing the scale and diversity of the dataset. This process enables the construction of a large collection of trajectories while relying on only a limited number of manually recorded demonstrations.

After data generation, the outputs produced by the official Isaac Lab scripts are reorganized and redirected into a unified episodic format that conforms to the dataset structure. This reorganization step preserves the original trajectory content while making the data compatible with our world model learning and downstream reinforcement learning pipelines.

For downstream reinforcement learning, we formulate a sparse episodic control task on top of the stacking environment. The reward is defined based on the final stable height of the stacked objects, measured after all object velocities have decayed to zero. This reward is provided once per episode and directly reflects task completion quality.

\subsection{AdaManip \texttt{Articulated} Task}

We construct an articulated-object manipulation dataset based on AdaManip~\citep{Wang2025AdaManipAA}. 
The dataset covers nine categories of articulated objects, with a total of 210 asset instances. Each asset is represented as a structured articulated object, consisting of geometry files together with an articulation specification that defines movable links, joints, and task-relevant operable parts. The assets span diverse mechanisms, including rotating and sliding caps, lockable handles, push-or-rotate controls, and switch-contact interactions. 
Table~\ref{tab:adamanip_assets} summarizes the asset categories used in our experiments.

\begin{table}[t]
\centering
\small
\caption{
\textbf{Articulated-object assets used in the AdaManip task.}
We use nine categories with different articulated mechanisms and operable parts.
}
\label{tab:adamanip_assets}
\setlength\tabcolsep{4.5pt}
\renewcommand{\arraystretch}{1.15}
\begin{tabular}{lcl}
\toprule
\textbf{Category} & \textbf{\# Instances} & \textbf{Mechanism / Operable Parts} \\
\midrule
Bottle          & 42 & cap; rotate-and-slide interaction \\
Pen             & 26 & cap; rotate-and-slide interaction \\
Coffee Machine  & 17 & portafilter; rotate-and-slide interaction \\
Window          & 18 & handle; lock and random rotation direction \\
Pressure Cooker & 18 & handle / lid; rotate-and-slide interaction \\
Lamp            & 24 & button / knob; push-or-rotate ambiguity \\
Door            & 31 & handle; lock and random rotation direction \\
Safe            & 18 & door / knob; lock and switch-contact interaction \\
Microwave       & 16 & door / button; lock and switch-contact interaction \\
\bottomrule
\end{tabular}
\end{table}

For data collection, we follow the policy and data-generation scripts provided by AdaManip. 
For each asset instance in every object category, we collect 20 episodes, and each episode contains 301 simulation steps. 
This results in 4,200 articulated-object manipulation episodes in total. 
Each episode records the executed robot actions, structured robot and object states, articulated joint states, static asset descriptors, and rendered visual observations, and is converted into the same episodic format used by the other tasks.

During trajectory generation, the camera pose and the initial robot state are kept fixed across episodes to ensure a consistent observation and control setup. 
To introduce controlled variation, we randomize the initial pose of the manipulated articulated object: its position is perturbed within $\pm 5$ cm, and its orientation angle is perturbed within $\pm 10^\circ$. 
This procedure preserves the underlying articulated mechanism of each asset while providing diverse initial configurations for learning object-centric articulated dynamics.

\subsection{Physion \texttt{Drape} Task}

We adopted Physion's point cloud dataset for arbitrarily shaped, deformable object dynamics. The Drape scene in Physion includes both irregularly shaped objects and a deformable cloth. At the start, a cloth is placed flat in the air, while several objects are placed on the floor. The objects include balls, cuboids, cones, cylinders, lamps, chairs, and so on. Then, the cloth in the air falls down, draping over the objects on the floor. During the scene, the cloth moves and deforms according to the objects' shapes, which requires \model{}-Dynamics to learn the deformation dynamics. The Physion dataset does not cover physical parameters or actions.






\subsection{MuJoCo \texttt{Cable Lift} Task}
\label{app:cable_data}

We construct an episodic trajectory dataset for a deformable-cable manipulation task in MuJoCo and evaluate the resulting pipeline on a real AgileX Cobot Magic dual-arm platform. In both settings, two robot arms grasp the endpoints of a cable and move within a shared vertical plane. The cable is represented by 25 ordered nodes, and the two interior nodes at indices 8 and 16 serve as task keypoints. Each episode specifies a target for each keypoint and uses a fixed 32-step control horizon. The action at each step is a four-dimensional planar end-effector command, with two dimensions assigned to each arm.

To model cables with different physical properties, the MuJoCo cable model exposes parameters including bending stiffness, bending-constraint iterations, elastic recovery strength, and damping, together with the cable geometry. These parameters determine how the cable bends, relaxes, and responds to the coordinated motion of the two arms. To reduce the sim-to-real gap, the simulated scene uses the same robot arrangement and matched camera intrinsics and extrinsics as the Cobot Magic setup. The initial robot configuration, cable geometry, and control protocol are fixed across episodes, while the cable parameters and target locations vary across the dataset. During data generation, the target is selected from a probe rollout under the current cable condition, and the probe is then replayed to create a reference trajectory. A second trajectory is generated with independently sampled actions under the same cable condition and target, yielding a paired counterfactual trajectory. The two roles are retained for paired analysis and are kept in the same data split.

Each episode records the normalized action sequence, aligned RGB and depth observations, robot poses, cable-node positions, and static descriptors including the cable parameters and target locations. The cable keypoints are derived from nodes 8 and 16 when needed rather than stored as a separate state field. The data therefore provide both structured state supervision for \model{}-Dynamics and image supervision for \model{}-Vision, while retaining the physical conditions needed for system-identification experiments.

We partition the cable conditions into training, in-distribution (ID), and out-of-distribution (OOD) subsets. ID conditions use cable parameters within the range observed during training but held out from the training trajectories, whereas OOD conditions use parameter combinations outside that range. Episodes from the same paired group are assigned to a single split to prevent leakage. We do not filter trajectories by a binary success label during data collection; continuous target distances are retained for analysis and downstream control experiments.

\section{Implementation Details}

\subsection{\model{}-Dynamics}
\label{app:dynamics}

We train the dynamics module using a two-stage procedure to balance one-step prediction accuracy and long-horizon rollout stability.
In the first stage, the model is optimized with teacher forcing, where ground-truth physical states are provided as inputs at every timestep.
This stage focuses on learning accurate single-step state transitions while avoiding early error accumulation.
In the second stage, we fine-tune the model under autoregressive rollouts by gradually replacing ground-truth inputs with the model’s own predictions.
The transition from teacher-forced training to fully autoregressive execution is controlled by an autoregressive ratio that increases over training epochs following a cosine annealing schedule, improving robustness to compounding errors.

The dynamics model predicts explicit physical states for the robot and all manipulated objects.
The robot state consists of position, linear velocity, and orientation represented as a flattened rotation matrix.
Each object state includes position, linear velocity, and orientation represented as a unit quaternion.

We supervise state predictions using a composite state loss, where robot and object losses are computed separately:
\begin{equation}
\mathcal{L}_{\text{state}} = \mathcal{L}_{\text{robot}} + \mathcal{L}_{\text{object}}.
\end{equation}

For translational position and velocity, we apply a Smooth-$\ell_1$ loss.
For rotational components, we use geometry-aware distance metrics:
The robot orientation loss is defined as the geodesic distance between predicted and ground-truth rotation matrices on $\mathrm{SO}(3)$,
while object orientations are supervised using a normalized quaternion distance.
Object losses are first averaged over objects and then over the batch.

During evaluation, we report a trajectory error (TrajErr) using the same state loss formulation.
Specifically, an inference module maps generated video sequences to explicit physical state trajectories, which are then compared against ground-truth trajectories using $\mathcal{L}_{\text{state}}$.
This metric directly measures long-horizon physical consistency of model rollouts and is aligned with the training objective of the dynamics module.

Model parameters are optimized using AdamW with a cosine-decayed learning rate schedule across all training stages.

\subsection{\model{}-Vision}

\begin{figure}[t]
    \centering
    \includegraphics[width=\linewidth]{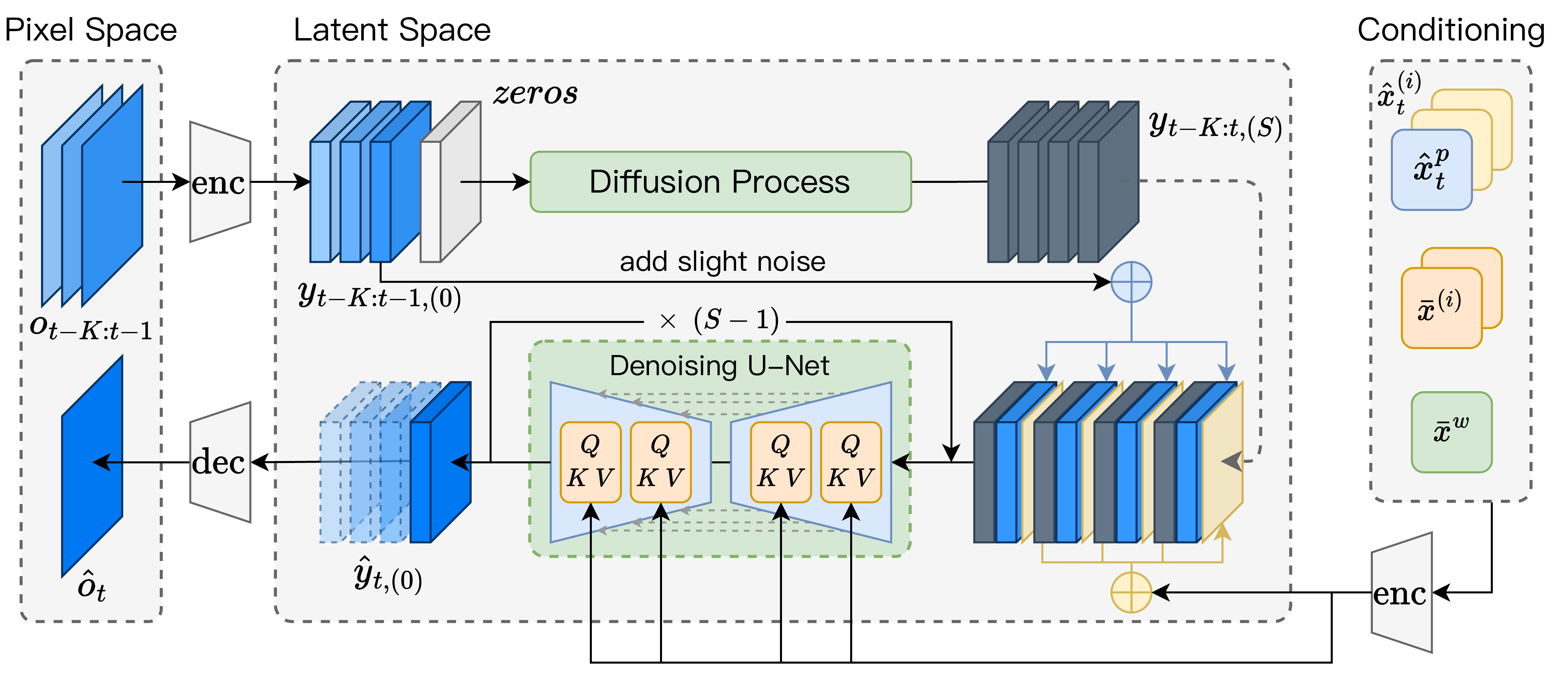}
    \caption{\textbf{\model{}-Vision}. 
    The denoising process is grounded in predicted physical states through spatial condition maps and object tokens, with context frames providing supplementary visual details.}
    \label{fig:vision}
\end{figure}

\model{}-Vision is trained as a latent diffusion model that generates visual observations conditioned on predicted physical states and scene descriptors.
In this section, we focus on training objectives and data sampling strategies that improve robustness under different rollout conditions.

We adopt the standard diffusion denoising objective in the latent space.
Each target frame is first encoded into a latent representation using a VAE encoder, and a forward diffusion process is applied to obtain noisy latents.
The denoising network is trained to recover the clean latent conditioned on the predicted physical state and available context frame latents.

To improve robustness to varying rollout settings, we randomize the number of context frames during training.
Specifically, for each training sample, we uniformly sample the context length $K \in \{0,1,\dots,6\}$ and condition the model on the corresponding past frames.
Importantly, we include the case $K=0$, which explicitly trains the model to reconstruct images solely from the given physical state without relying on visual context.
This encourages stable state-to-image rendering when context frames are absent or unreliable.

In addition to varying the context length, we randomize how context frames are selected.
With a certain probability, we use temporally consecutive context frames and predict the next consecutive frame.
Otherwise, we randomly sample $(K{+}1)$ frames from a full episode at different timestamps, sort them by temporal order, and use the first $K$ frames as context to predict the $(K{+}1)$-th frame.
By increasing the temporal gaps between context frames, this stochastic sampling strategy reduces over-reliance on short-term visual continuity and encourages the model to generate observations that are consistent with the provided physical state.

\subsection{Extension to Complex Objects}
\label{app:complex_objects}

We extend \model{} to more complicated object types by adapting the construction of static scene descriptors and physical states while preserving the same object-centric dynamics formulation. 
For articulated objects, the key challenge is to encode structured multi-part geometry together with articulation priors. 
For deformable objects, the main difficulty is that object geometry evolves over time and therefore cannot be treated purely as a static asset attribute.

\paragraph{Articulated objects.}
For articulated objects, we represent each asset as a collection of semantic part-level tokens rather than encoding the entire object as a single rigid entity. 
Concretely, we first use the URDF specification to determine the part decomposition of the asset, the correspondence between each part and its associated mesh, and the articulation metadata, including joint type, axis, limits, and parent-child relations. 
For each part, we then recover its triangle mesh, transform it into the corresponding part-local visual frame according to the URDF scale and visual origin, and uniformly sample surface point clouds with normals. 
The sampled point cloud is normalized and passed through a pretrained Sonata~\citep{Wu2025SonataSL}/PTV3~\citep{Wu2023PointTV} encoder, and the resulting point features are deterministically pooled into a fixed-dimensional latent representation. 
The final part-level latent, together with the associated geometric and joint metadata, is stored as the asset attribute of that part. 
In this way, an articulated object is converted into a structured set of part-level descriptors that capture both geometry and articulation priors, and can be directly incorporated into the static scene descriptor of \model{}-Dynamics. 

\paragraph{Deformable objects.}
For deformable objects, we adopt a different treatment because the object shape is inherently time-dependent. 
In the Drape~\citep{Bear2021PhysionEP} environment, the static objects on the floor remain fixed throughout the episode and, for simplicity, are treated as one large static object, while the cloth is modeled as a separate deformable object. 
We extract point-cloud-based embeddings for both the static scene object and the deformable cloth using Point Transformer V3~\citep{Wu2023PointTV}, yielding object-level features whose dimensionality depends only on the number of objects rather than the number of sampled points. 
This matches the object-centric encoder design of \model{}-Dynamics and allows deformable scenes to be represented within the same structured input interface. 
Unlike rigid or articulated assets, however, the shape of the cloth changes over time and therefore cannot be encoded only as a static attribute. 
Instead, the corresponding shape features are treated as part of the physical state and updated along the temporal rollout. 
This design enables \model{} to naturally extend from rigid-body manipulation to deformable dynamics while maintaining a unified object-centric representation.

\section{More Experimental Results}
\label{app:more_exprs}

\subsection{\model{}-Dynamics Evaluation}
\label{app:dynamics_eval}

\begin{figure}[t]
    \centering
    \includegraphics[width=0.99\columnwidth]{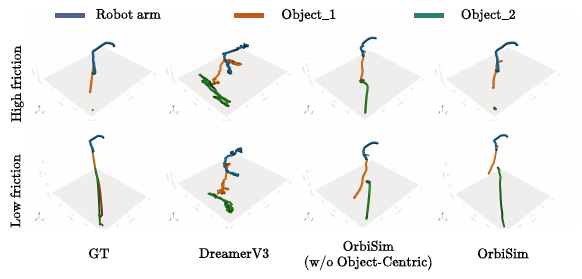}
    \caption{\textbf{Physical state trajectories by \model{}-Dynamics.} The autoregressive rollouts demonstrate high sensitivity to physical parameters, accurately distinguishing between low- and high-friction regimes while maintaining smoothness over long horizons.
    }
    \label{fig:vis_state}
\end{figure}

We evaluate the state-space generative performance of \model{}-Dynamics by benchmarking it against DreamerV3-Torch on the robosuite offline dataset. As illustrated in Figure \ref{fig:vis_state}, our dynamics core effectively models complex state transitions while demonstrating a high degree of sensitivity to underlying physical configurations.

Specifically, the results show that \model{}-Dynamics accurately distinguishes between low- and high-friction regimes, adjusting object trajectories accordingly while maintaining predictive smoothness over long-horizon rollouts. This responsiveness to varying object-table friction coefficients confirms the model's capacity to ground its learned dynamics in explicit physical parameters, a critical requirement for robust system identification and transferable control.

\subsection{Workflow Analysis}
\label{app:worlflow}

To disentangle the effects of grounding, autoregressive feedback, and visual context, we evaluate four distinct execution workflows:
$\bullet$
\emph{GT-first} initializes rollouts with ground-truth states and observations, after which \model{} autoregressively predicts both future states and images.
$\bullet$
\emph{Recon-first} initializes rollouts from a ground-truth state only, generating the first observation from the state without visual context and then rolling out autoregressively.
$\bullet$
\emph{GT-state} replaces autoregressive state predictions with ground-truth state sequences, evaluating visual rendering under perfectly grounded dynamics.
$\bullet$
\emph{Recon-all} predicts each observation independently from the corresponding physical state, without using any temporal visual context.

As shown in \tabref{tab:wm_push1}, a compelling trade-off is observed in the \textbf{Recon-all} variant. It attains substantially higher PSNR due to its bias toward per-frame reconstruction accuracy; however, its degradation in perceptual similarity (LPIPS) and temporal coherence (FVD) indicates a lack of long-horizon visual consistency. 
In contrast, \textbf{GT-first} and \textbf{GT-state} strike a superior balance, demonstrating that \model{} effectively utilizes temporal context to stabilize long-horizon rollouts.

\begin{table*}[t]
\centering
\small
\caption{
    \textbf{Analysis of \model{} rollout variants.}
    (i) \textbf{GT-first:} Autoregressive rollout given true $\{o_0,x_0,\bar{x}\}$;
    (ii) \textbf{GT-state:} Uses true state sequences throughout to isolate rendering performance;
    (iii) \textbf{Recon-first:} Autoregressive rollout only from true initial state;
    (iv) Recon-all: Evaluates per-frame state-to-pixel fidelity by excluding all temporal visual context during generation, \ie $p_\phi^{\text{vis}}(o_t \mid \hat{x}_t, \bar{x})$.
    \textbf{TrajErr}: Measures the discrepancy between \textit{inferred physical states from generated videos} (see Sec.~\ref{sec:real2sim}) and the ground truth.
}
\label{tab:wm_push1}
\setlength\tabcolsep{5pt}
\renewcommand{\arraystretch}{1.15}
\begin{tabular}{lcccccccc}
\toprule
\textbf{Method}
& \textbf{PSNR10} $\uparrow$
& \textbf{PSNR100} $\uparrow$
& \textbf{LPIPS10} $\downarrow$
& \textbf{LPIPS100} $\downarrow$
& \textbf{FVD} $\downarrow$
& \textbf{TrajErr} $\downarrow$\\
\midrule
\textbf{\model{} (GT-first)}
&\underline{26.7105}  &21.8269
&\textbf{0.1078}  &\underline{0.1428}  &\textbf{533.9} & 0.4468 \\
\textbf{\model{} (GT-state)}
&\textbf{26.7253} &\underline{20.2680}
&\underline{0.1087}  &\textbf{0.1388}  &\underline{537.3} & \underline{0.4322} \\
\textbf{\model{} (Recon-first)}
&24.6366  &19.3114
&0.1236  &0.1509  &553.6 & 0.4558 \\

\textbf{\model{} (Recon-all)}
&26.4910  &\textbf{23.3776}
&0.1279  &0.1628  &922.1 & \textbf{0.2927} \\
\bottomrule
\end{tabular}
\end{table*}

\subsection{Multi Objects / Multi Tasks}
\label{app:multi}


\begin{table}[h]
    \centering
    \small
    \caption{\textbf{Results on multi-object, multi-task datasets.} Single represents \model{} and AdaWorld trained on single-task robosuite \texttt{Push} dataset with up to 2 objects. Multi represents \model{} and AdaWorld trained on the extended \texttt{Push\_cubes} and \texttt{Pick\_and\_place} dataset with up to 4 objects. For data with more than 100 frames, the results end with ``100'' are from all frames.}
\setlength\tabcolsep{5pt}
    \begin{tabular}{lccccc}
        \toprule
        \textbf{Method}
        & \textbf{PSNR10} $\uparrow$
        & \textbf{PSNR100} $\uparrow$
        & \textbf{LPIPS10} $\downarrow$
        & \textbf{LPIPS100} $\downarrow$
        & \textbf{FVD} $\downarrow$ \\\midrule
        \textbf{OrbiSim (single)} & 26.7105 & 19.9819 & 0.1078 & 0.1428 & 533.9 \\
        \textbf{AdaWorld (single)} & 26.6647 & 12.8346 & 0.1183 & 0.3482 & 1305.8 \\
        \textbf{OrbiSim (multi)} & 24.8676 & 18.4515 & 0.1210 & 0.1959 & 863.1 \\
        \textbf{AdaWorld (multi)} & 24.4685 & 14.3235 & 0.1332 & 0.2661 & 1292 \\\bottomrule
    \end{tabular}
    \label{tab:multi_tasks_main}
\end{table}

We conduct a scaling study focusing on predictive fidelity and memory overhead in multi-task multi-object scenarios.

For predictive fidelity, we evaluate \model{} and AdaWorld baseline in the extended multi-object robosuite dataset, namely \texttt{Push\_cubes} and \texttt{Pick\_and\_place}. Dataset configuration and collection details are discussed in Appendix~\ref{app:robosuite_multi}.
These task scenes contain more than 2 objects with a wide range of task settings. These tasks include not only cube pushing, hitting, and friction, but also objects falling, stacking, and robot gripper picking. Thus, these two categories are primitive and fundamental tasks for most robot arm tasks, requiring \model{} to learn 3-dimensional object positioning, interaction, and gripping.

\begin{wrapfigure}{r}{0.4\textwidth} 
\vspace{-6pt}
    \centering
    \includegraphics[width=0.4\textwidth]{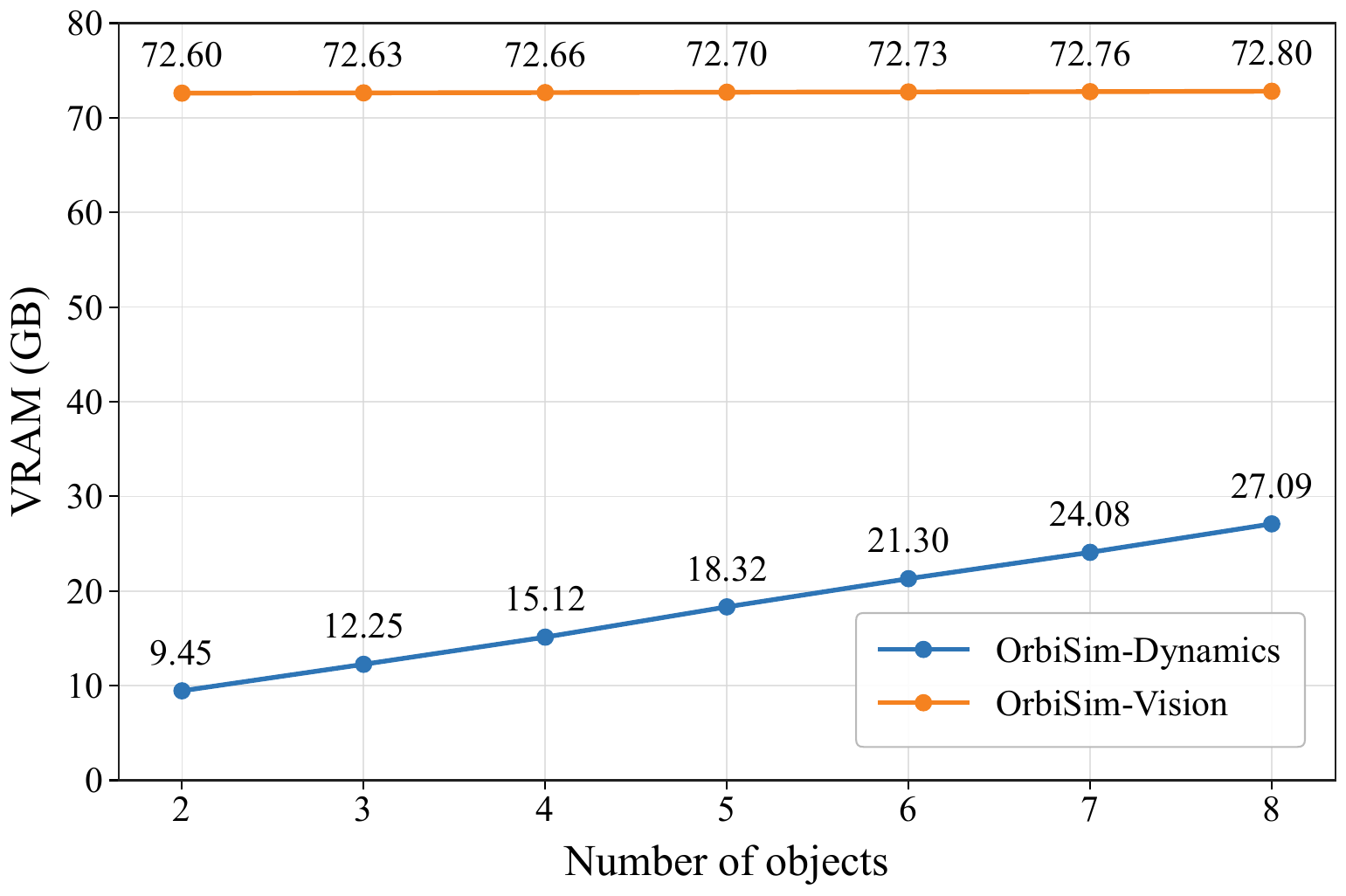}
    \caption{\textbf{Memory overhead results from object number scaling.}}
    \label{fig:mem_scale}
\vspace{-16pt}
\end{wrapfigure}

The main results of multi-object and multi-task world prediction are in \tabref{tab:multi_tasks_main}. Although it experiences performance degradation in multi-object, multi-task scenarios, \model{} still achieves higher visual quality and consistency scores.

Additionally, we measured the training memory overhead as the number of objects increased from 2 to 8. The results are shown in \figref{fig:mem_scale}.
As shown in the results, VRAM usage grows approximately linearly. The VRAM usage of \model{}-Dynamics is nearly proportional to the number of objects, while \model{}-Vision's VRAM growth is tiny compared to its original overhead.
This demonstrates that the Transformer-based coupling remains computationally manageable for medium-scale scenes.

\subsection{Robustness}
\label{app:robustness}


We conduct experiments on noisy states to quantify \model{}'s robustness. 
We set two levels of Gaussian noise on physical and visual states:
\begin{itemize}[leftmargin=*]
\item Moderate Noise: Small-scale perturbations to physical states. 
\item Strong Noise: Significant noise levels designed to simulate high-error visual state estimation.
\end{itemize}

\begin{table}[h]
    \centering
    \small
\vspace{-8pt}
    \caption{\textbf{Results on noisy states rollout.}}
\setlength\tabcolsep{5pt}
\renewcommand{\arraystretch}{1.15}
    \begin{tabular}{lccc}
    \toprule
        \textbf{Method} 
        & \textbf{PSNR100 $\uparrow$}
        & \textbf{LPIPS100 $\downarrow$}
        & \textbf{FVD $\downarrow$} \\\midrule
        \textbf{\model{}} & 19.9819 & 0.1428 &	533.9 \\
        \textbf{\model{} with moderate noise} &	20.2508 &	0.1309 &	587.14 \\
        \textbf{\model{} with strong noise} &	20.1911 &	0.1337 &	606.98 \\\bottomrule
    \end{tabular}
    \label{tab:noisy_states}
\end{table}

As shown in \tabref{tab:noisy_states}, \model{} exhibits substantial resilience to noise on both physical and visual states. Even under high levels of visual noise, \model{} remains highly robust, exhibiting low degradation in visual generation.
This demonstrates that \model{}'s framework can tolerate imperfection in physical and visual states, which is promising for potential Real-to-Sim scenarios even where ground-truth states are unavailable.

\subsection{Imitation Learning Experiments on the Isaac Lab \texttt{Stack} Task}
\label{app:imitation_learning}

\begin{table}[t]
\centering
\small
\caption{\textbf{Imitation learning performance on the Isaac Lab \texttt{Stack} task.}
The imitation learning baseline is implemented using the official Isaac Lab
demonstration-based example without reinforcement learning.
Results highlight the sensitivity of pure imitation learning to
distributional shifts in long-horizon manipulation tasks.}
\label{tab:il_stack}
\setlength\tabcolsep{15pt}
\begin{tabular}{lccc}
\toprule
\textbf{Method} & \textbf{Input}
& \textbf{Success (\%)} \\
\midrule
Isaac Lab-IL (Vision)  & RGB + State   &  78.53 \\
Isaac Lab-IL (Noise)   & Noise + State &  0.00  \\
Isaac Lab-IL (State)   & State         &  78.00 \\
\bottomrule
\end{tabular}
\end{table}

We further conduct imitation learning experiments on the Isaac Lab \texttt{Stack} task, which is substantially more challenging than the pushing task due to its longer horizons, multi-stage object interactions, and stricter requirements on precise state coordination. The stacking task requires not only reaching intermediate configurations, but also maintaining stability under contact-rich dynamics, making it particularly sensitive to compounding errors.

As shown in Table~\ref{tab:il_stack}, the imitation learning baseline
(\textbf{Isaac Lab-IL}) can achieve reasonable performance when evaluated
under in-distribution conditions that closely match the demonstration data.
However, its performance degrades catastrophically under out-of-distribution
(OOD) settings, such as variations in object configurations or interaction dynamics.
This failure mode highlights a fundamental limitation of pure imitation learning:
without an explicit mechanism for modeling environment dynamics or long-horizon
credit assignment, small deviations from the demonstration manifold can rapidly
accumulate and lead to irreversible task failure.

These results motivate the use of reinforcement learning, particularly when
augmented with learned world models, to enable robust long-horizon decision making
and recovery under distributional shifts.

\subsection{Orbisim-Dynamics Compared to Newton Simulator}
\label{app:newton}

\begin{wrapfigure}{r}{0.54\textwidth}
\vspace{-8pt}
    \centering
    \includegraphics[width=\linewidth]{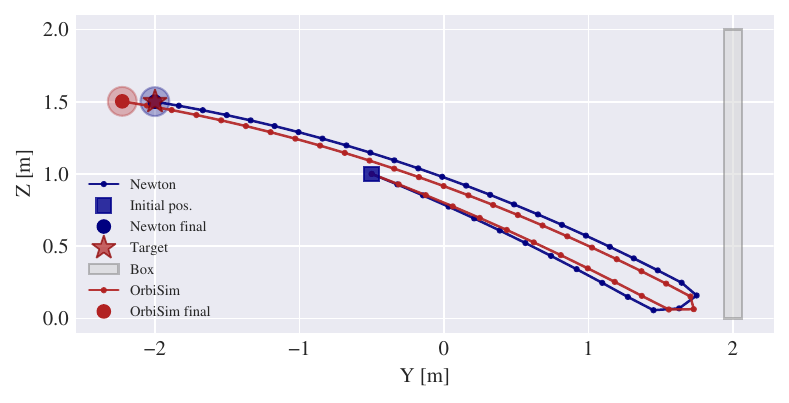}
    \caption{\textbf{Rollout trajectory in Newton simulator.} Both trajectories initialized with velocity optimized in Newton and \model{}-Dynamcics are shown. \model{}-Dynamics' result shows a tolerable difference.}
    \label{fig:newton_traj}
\vspace{-8pt}
\end{wrapfigure}

To prove the effectiveness of analytical gradients of \model{}, we conducted experiments comparing \model{} to differentiable physics simulators. 
We adopted NVIDIA Newton as a baseline, with its diffsim tasks simulated in Warp differentiable simulator. 
The \texttt{diffsim\_ball} task is chosen. At the start, a ball is in the air with an initial velocity. The task requires optimization over initial velocity, making the ball bounce on the floor and a static wall, and finally reach a target position. No robot arms or actions are included in this task.
In newton simulator, the optimization is achieved by the raw gradient descent method, utilizing the gradient of the squared target position error through the simulator. 

For optimization in \model{}, we first collect a dataset by rollout in Newton with randomly-sampled initial velocity. A total of 500 episodes are collected for \model{}-Dynamics training. 
After training, we optimized initial velocity through analytical gradients from \model{}-Dynamics with the same method as Newton.

\begin{figure}[htbp]
    \centering
    \includegraphics[width=\linewidth]{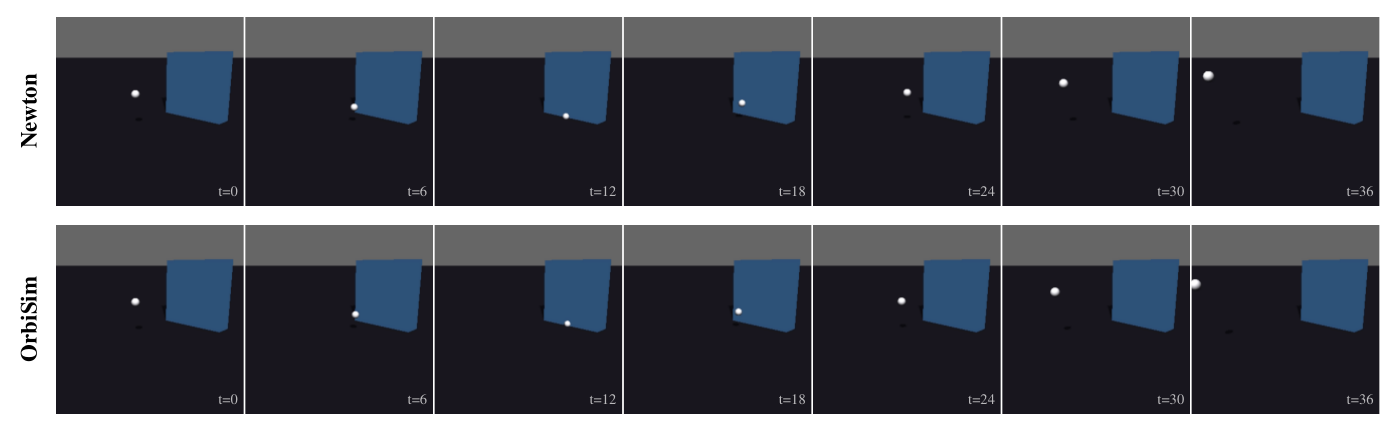}
    \caption{\textbf{Visual rollout in Newton simulator.} The two rows represent rollouts with initial velocity optimized with Newton warp and \model{}-Dynamics, respectively.}
    \label{fig:newton_vis}
\end{figure}

With a learning rate of 0.8, gradient descent in Newton simulator reaches zero loss within 60 steps. We conduct gradient descent with the same learning rate in \model{}-Dynamics, and the loss value is reduced to less than 1e-4 within 800 steps. 

The visual results are shown in \figref{fig:newton_vis} and \figref{fig:newton_traj}. 
The initial velocity optimized from newton is $[0.0, 10.639, -4.282]$, while the velocity optimized from \model{}-Dynamics is $[-0.07, 11.218, -4.238]$. A tolerable difference is shown in the velocity and rollout trajectories.

\model{} shows a slight error in dynamics estimation accuracy compared to simulators. However, differentiable simulators such as Newton DiffSim only support limited scenarios, while \model{} supports a wider variety of object types, scene configurations, and physics settings.

\subsection{More Downstream Reinforcement Learning Experiments}
\label{app:rl}


\begin{figure*}[t]
    \centering
    \begin{minipage}[t]{0.9\textwidth}
        \centering
        \includegraphics[width=\linewidth]{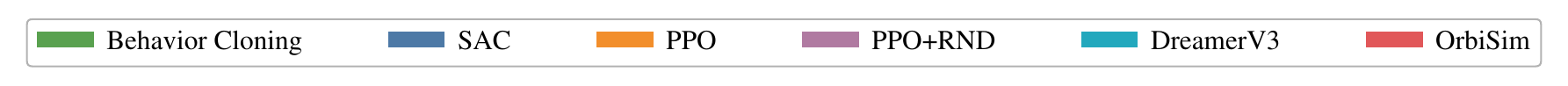}
    \end{minipage} \\
    \begin{minipage}[t]{0.24\textwidth}
        \centering
        \includegraphics[width=\linewidth]{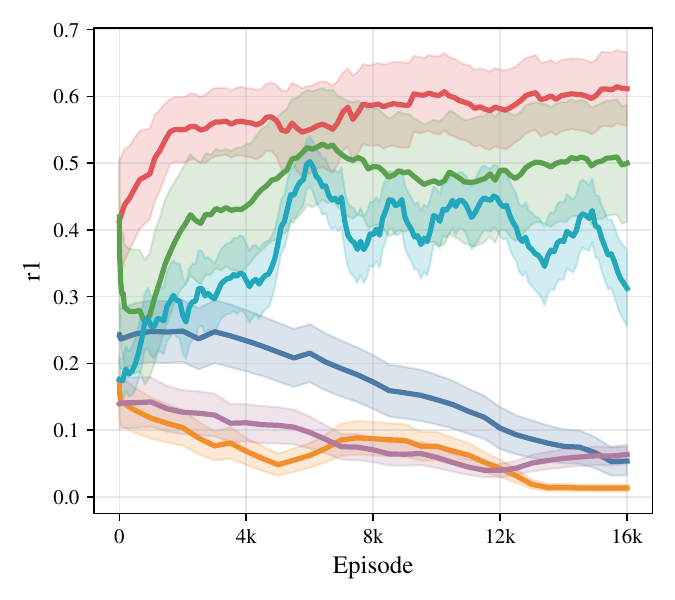}
        \makebox[\linewidth]{(a) $r_1$}
    \end{minipage}
    \begin{minipage}[t]{0.24\textwidth}
        \centering
        \includegraphics[width=\linewidth]{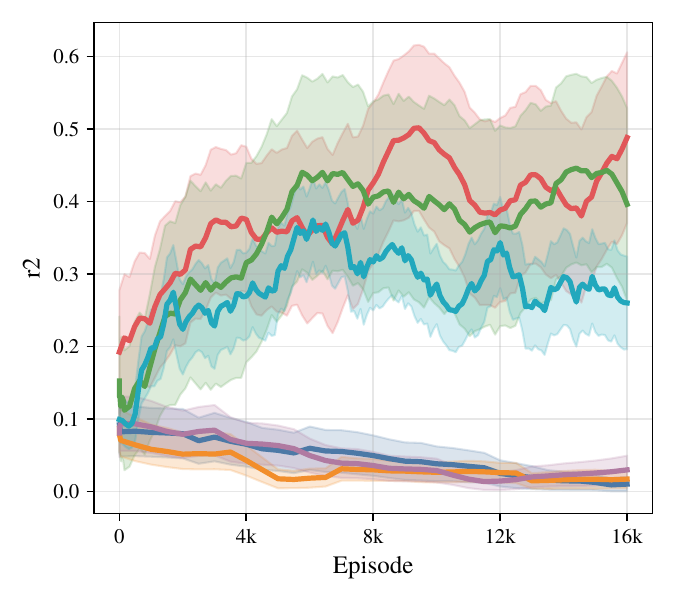}
        \makebox[\linewidth]{(b) $r_2$}
    \end{minipage}
    \begin{minipage}[t]{0.24\textwidth}
        \centering
        \includegraphics[width=\linewidth]{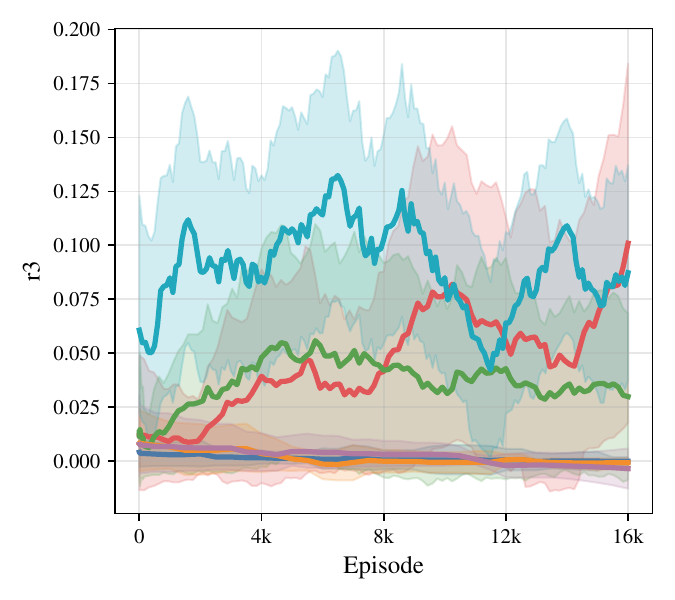}
        \makebox[\linewidth]{(c) $r_3$}
    \end{minipage}
    \begin{minipage}[t]{0.24\textwidth}
        \centering
        \includegraphics[width=\linewidth]{fig/all_rewards_smooth_total.pdf}
        \makebox[\linewidth]{(d) Total reward}
    \end{minipage}
    \vspace{-3pt}
    \caption{\textbf{Training curves on the robosuite \texttt{Push} task.}
    The x-axis denotes training episodes and the y-axis denotes normalized episode rewards defined in Sec.~\ref{sec:rl_eval} and Eq. \eqref{eq:rl_rewards}, namely $r_1$, $r_2$, $r_3$, and the total reward. See reward definition in the text.
    }
    \label{fig:rl_training_curves_detailed}
\end{figure*}

\begin{table}[t]
    \centering
    \caption{
    \textbf{Success rate on the robosuite \texttt{Push} task.} We evaluate \model{}, model-based baseline (DreamerV3), Behavior Cloning (BC) and model-free baselines (SAC, PPO, PPO+RND) in the robosuite environment. Success is achieved if the second cube moves more than 5cm.
    }
    \begin{tabularx}{\textwidth}{c  *{6}{>{\centering\arraybackslash}X}}
        \toprule
        {Method} & \textbf{\model{}} & DreamerV3 & BC & PPO+RND & PPO & SAC \\\midrule
        {Success Rate(\%)} & \textbf{42.71} & 25.00 & 19.79 & 2.08 & 1.04 & 0.00 \\\bottomrule
    \end{tabularx}
    \label{tab:rl_success}
\end{table}

In the robosuite environment, we define three normalized terminal rewards based on distance reduction: $r_1$ encourages approaching the first cube, $r_2$ promotes contact between the two cubes, and $r_3$ measures task progress by pushing the second cube toward the table boundary. Each reward is computed as the relative decrease from its initial distance. For $r_1$ and $r_2$, the minimum distance over the episode is used to capture transient interactions, while $r_3$ uses the final distance to reflect task completion. Specifically, for an episode of $T$ steps, let $d_{\text{r, c1}, t}$ denote the distance between the robot arm (r) and the first cube (c1) at timestep $t$, and the rewards are defined as:
\begin{equation}
    r_1 = 1 - \frac{\min_{t=1}^{T} d_{\text{r, c1}, t}}{d_{\text{r, c1}, 0}}, \quad
    r_2 = 1 - \frac{\min_{t=1}^{T} d_{\text{c1, c2}, t}}{d_{\text{c1, c2}, 0}}, \quad
    r_3 = 1 - \frac{x_{\text{table\_border}} - x_{\text{c2}, T}}{x_{\text{table\_border}} - x_{\text{c2}, 0}}.
\label{eq:rl_rewards}
\end{equation}
During training, these three rewards are provided only at episode termination, resulting in sparse learning signals that are challenging for conventional policy optimization methods. We also use these three rewards to evaluate task completion performance across different methods. Success is defined in robosuite \texttt{Push} task if the second cube has moved more than 5cm during rollout. 

The detailed training curves are provided in \figref{fig:rl_training_curves_detailed}. The success rate comparison is provided in \tabref{tab:rl_success}. \model{} performs higher task completion rewards and success rate than all baselines.

To qualitatively evaluate policy performance on the downstream task, we visualize representative rollouts produced by policies trained using 
model-free RL, model-based RL and Behavior Cloning (BC).
For model-free RL, we include Soft Actor-Critic (SAC), Proximal Policy Optimization (PPO) and PPO with Random Network Distillation (PPO+RND). 
For model-based RL, we adopt DreamerV3 as a baseline.

\begin{figure}[htbp]
    \centering
    \includegraphics[width=\linewidth]{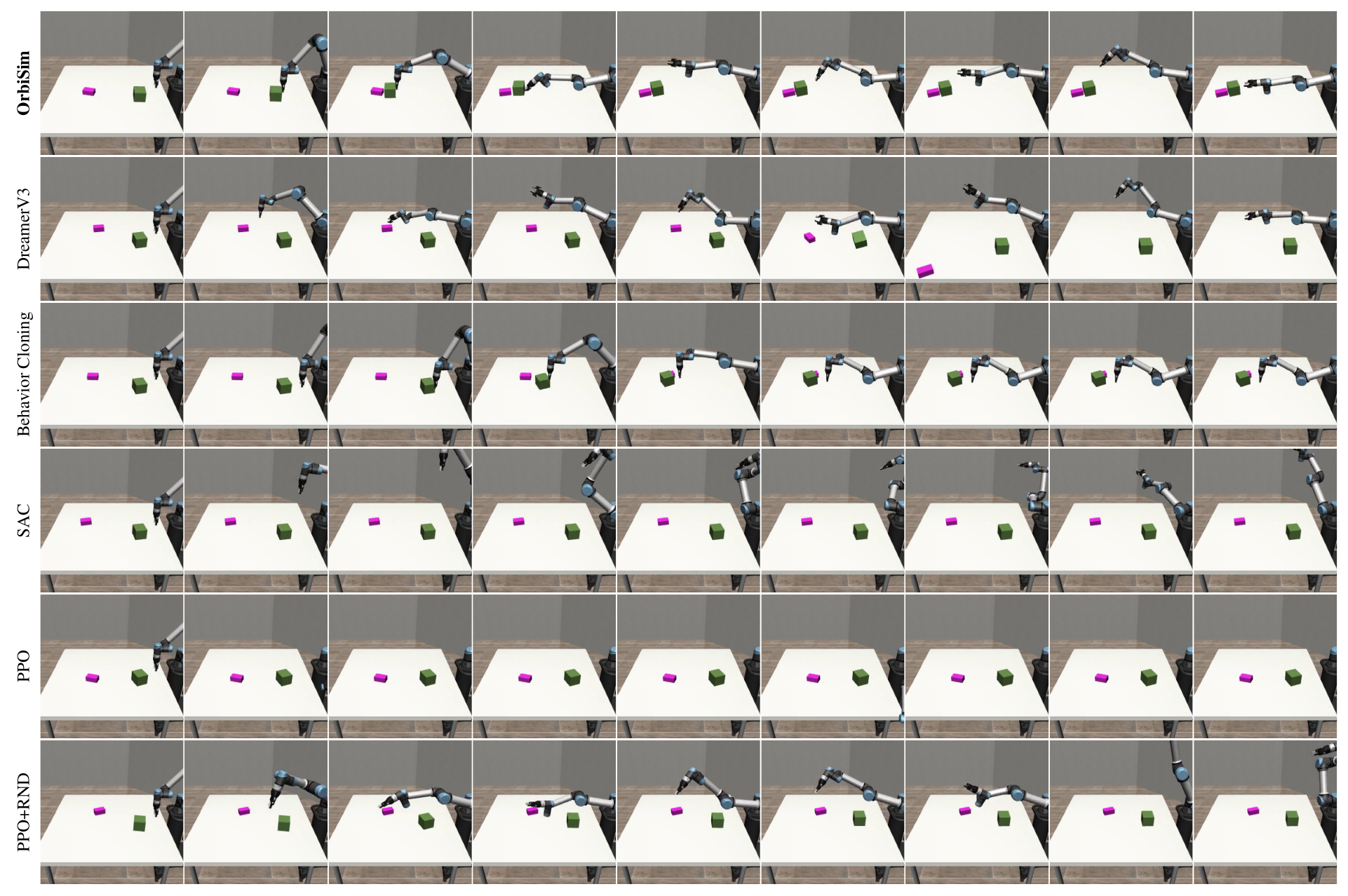}
    \caption{\textbf{Reinforcement learning results on robosuite \texttt{Push} task.} We show video rollouts in robosuite simulator of \model{} model-based policies trained with \model{} dynamics, DreamerV3 model-based policies trained with DreamerV3 prediction, behavior cloning trained with expert policies, and model-free baselines trained with robosuite environment.}
    \label{fig:rl_result_video}
\end{figure}

As shown in Fig.~\ref{fig:rl_result_video}, policies trained with model-free RL fail to learn effective behaviors for the downstream task, resulting in disordered motions and unsuccessful task completion. 
DreamerV3 policies perform more effectively but still fail in stability and generalization in unseen distributions.
Behavior Cloning (BC) policies, while able to reproduce demonstrations to some extent, exhibit unstable performance and frequently deviate from desired behaviors due to the accumulation of prediction errors over long horizons. 
In contrast, our method consistently produces coherent and goal-directed behaviors, enabling successful task execution across different scenarios. These visual comparisons demonstrate that our approach provides more reliable policy performance on the downstream task compared with both reinforcement learning and imitation learning baselines.

\subsection{Simulation and Real-World Results on MuJoCo \texttt{Cable Lift} Task}
\label{app:cable}

\paragraph{Parameter identification results.}
The simulation study evaluates the complete alternating procedure for physical-parameter identification and action optimization. For each fixed test setting, we initialize the parameter estimate and action sequence, identify the cable properties while holding the actions fixed, and then optimize the action sequence with the updated differentiable dynamics model. We report results separately for in-distribution (ID) settings, whose cable parameters are within the training range, and out-of-distribution (OOD) settings, whose parameters lie outside that range. The state-based route uses reference state trajectories for SysID, whereas the vision-based route uses discrepancies between predicted and reference image trajectories. In both routes, the identified parameters are subsequently used for state-space action optimization.

Table~\ref{tab:cable-sysid} reports the final mean absolute error of the four cable parameters over each split. The parameters are the bending stiffness coefficient (\texttt{bend}), the number of bending-constraint iterations (\texttt{bend\_passes}), the elastic recovery strength (\texttt{recovery}), and the damping coefficient (\texttt{damping}). The state-based route achieves lower errors for all four parameters on both ID and OOD settings, as expected from its access to privileged state supervision. The vision-based route also recovers the parameter setting consistently, with a larger residual error in OOD settings, especially for the recovery and damping terms. These estimates are nevertheless sufficient to support the subsequent action optimization, as reflected by the reduction in terminal keypoint-to-target distance reported in the main paper.

\begin{table}[t]
    \centering
    \small
    \caption{\textbf{Simulation SysID results on the \texttt{Cable Lift} task.} Entries are mean absolute errors of the final estimated cable parameters within each evaluation split. The \texttt{bend\_passes} column reports a discrete iteration-count error; the other columns use their respective simulator parameter units.}
    \label{tab:cable-sysid}
    \setlength{\tabcolsep}{5pt}
    \begin{tabular}{lcccc}
        \toprule[1.5pt]
        Setting & \texttt{bend} & \texttt{bend\_passes} & \texttt{recovery} & \texttt{damping} \\
        \midrule[1pt]
        State, ID   & 0.0025 & 0.10 & 0.0061 & 0.0093 \\
        State, OOD  & 0.0055 & 0.55 & 0.0238 & 0.0116 \\
        Vision, ID  & 0.0088 & 0.90 & 0.0185 & 0.0366 \\
        Vision, OOD & 0.0196 & 1.80 & 0.0778 & 0.0487 \\
        \bottomrule[1.5pt]
    \end{tabular}
\end{table}

The parameter errors should be interpreted together with the downstream task metric. Exact recovery of every simulator parameter is not required for successful control: the vision-based estimates can retain nonzero parameter error while still providing informative dynamics gradients and substantially improving the terminal keypoint distance. The gap between ID and OOD results also quantifies the additional difficulty of identifying physical conditions outside the training range.

\paragraph{Action execution results.}
Table~\ref{tab:cable-distance} compares the final sum of the two keypoint-to-target distances for the simulation and real-world evaluations. Simulation distances are measured in meters after MuJoCo replay and are reported separately for ID and OOD settings. The real-world metric is measured in pixels. The differentiable action-optimization routes substantially reduce the simulated terminal distance relative to SAC and $\pi_{0.5}$, with the state-based route giving the lowest residual error and the vision-based route remaining effective under both parameter splits. For the real-world comparison, the available $\pi_{0.5}$ and vision-based SysID results are reported in the corresponding rows; unavailable method--setting pairs are marked by dashes.

\begin{table}[t]
    \centering
    \small
    \caption{\textbf{Final keypoint-to-target distance on the \texttt{Cable Lift} task.} Simulation entries report the mean sum of the two keypoint-to-target distances after MuJoCo replay in meters, separately for ID and OOD settings. The real-world entry reports the mean distance in pixels. A dash indicates that the corresponding method was not evaluated in that setting.}
    \label{tab:cable-distance}
    \label{tab:cable-real}
    \setlength{\tabcolsep}{5pt}
    \begin{tabular}{lccc}
        \toprule[1.5pt]
        Method & \multicolumn{2}{c}{Simulation (m)} & Real world (px) \\
        \cmidrule(lr){2-3}
        & ID & OOD & \\
        \midrule[1pt]
        SAC & 0.4402 & 0.3678 & -- \\
        $\pi_{0.5}$ & 0.1124 & 0.0881 & 62.2 \\
        \model{} (state-based SysID) & 0.0122 & 0.0153 & -- \\
        \model{} (vision-based SysID) & 0.0240 & 0.0290 & 17.5 \\
        \bottomrule[1.5pt]
    \end{tabular}
\end{table}

The comparison highlights the distinction between the two evaluation settings. In simulation, both parameter-identification routes support direct action optimization and outperform the policy references under both ID and OOD conditions. The state-based route benefits from privileged state supervision, while the vision-based route achieves a larger but still small residual distance. In the real-world trials, the vision-based SysID pipeline reaches a substantially smaller pixel distance than $\pi_{0.5}$, showing that the differentiable rollout can provide useful control signals under the remaining sim-to-real gap.

\subsection{Computational Efficiency and Resource Usage}
\label{app:compute}

We provide additional computational statistics for representative components and baselines used in our experiments. 
Table~\ref{tab:compute_efficiency} reports the training throughput and GPU memory usage measured during downstream policy learning. 
All measurements are collected under the same simulator and training environment settings.

The results show that \model{} achieves substantially higher training throughput than latent world-model approaches such as DreamerV3 while maintaining moderate memory usage. 
Pure model-free baselines such as SAC achieve higher raw environment throughput, but fail to obtain competitive downstream control performance in our tasks. 
These observations support the practical benefit of structured end-to-end differentiability for efficient embodied policy optimization.

\begin{table}[h]
\centering
\small
\caption{\textbf{Training throughput and GPU memory usage of representative methods.}}
\label{tab:compute_efficiency}

\setlength{\tabcolsep}{15pt}
\begin{tabular}{lccc}
\toprule
Method & Steps/s & Device & GPU Memory \\
\midrule
\model{} dynamics & 29.41 & RTX 3080 & 372 MB \\
Behavior Cloning & 47.71 & RTX 3080 & 27.14 MB \\
DreamerV3 & 4.54 & RTX 4090D & 931 MB \\
SAC & 66.05 & RTX 3080 & 33.79 MB \\
\bottomrule
\end{tabular}

\end{table}

\end{document}

%% file: our_commands.tex
\usepackage{duckuments}
\usepackage{CJKutf8}
\usepackage{overpic}
\usepackage{pifont}
\usepackage{bbding}
\usepackage{comment}
\usepackage{float}
\usepackage{enumitem}

\definecolor{MyDarkBlue}{rgb}{0,0.5,1}
\definecolor{MyDarkGreen}{rgb}{0.02,0.6,0.02}
\definecolor{MyDarkRed}{rgb}{0.8,0.02,0.02}
\definecolor{MyDarkOrange}{rgb}{0.40,0.2,0.02}
\definecolor{MyYellow}{rgb}{1,0.55,0}
\definecolor{MyPurple}{RGB}{111,0,255}
\definecolor{MyRed}{rgb}{1.0,0.0,0.0}
\definecolor{MyGold}{rgb}{0.75,0.6,0.12}
\definecolor{MyDarkgray}{rgb}{0.66, 0.66, 0.66}
\definecolor{default}{RGB}{0,0,0}

\newcommand\eg{\textit{e.g., }}
\newcommand\ie{\textit{i.e., }}

\newcommand{\model}{OrbiSim} 
\newcommand{\figref}[1]{Figure~\ref{#1}} 
\newcommand{\tabref}[1]{Table~\ref{#1}} 

\newcommand{\mycheckmark}{{\textcolor{MyDarkGreen}{\checkmark}}}
\newcommand{\myxmark}{{\textcolor{MyDarkRed}{\ding{55}}}}